\newif\ifarxiv
\arxivtrue

\ifarxiv
    \documentclass{article}
    
    \usepackage{fullpage} 
    \usepackage{authblk}
    \usepackage{natbib}
    \usepackage{bbm}   \usepackage{amsmath,amssymb,amsthm,mathrsfs,pgf,tikz,caption,subcaption,mathtools,mathabx}
    \usepackage{algorithm, algorithmic}
    \usepackage[utf8]{inputenc}

    \usepackage[hyphens]{url}
    \usepackage[hidelinks]{hyperref}
    \hypersetup{breaklinks=true}
    \usepackage{graphicx} 
    \usepackage{xcolor}
    \usepackage{float}
\else
    \documentclass[manuscript,screen,review,anonymous]{acmart}
\fi

\newcommand{\cG}{\mathcal{G}}
\newcommand{\cH}{\mathcal{H}}

\newcommand{\D}{\mathcal{D}}
\newcommand{\X}{\mathcal{X}}
\newcommand{\Y}{\mathcal{Y}}
\newcommand{\Z}{\mathcal{Z}}
\newcommand{\Loss}{\mathcal{L}}
\newcommand{\mean}{\mathbb{E}}

\ifarxiv
\newtheorem{theorem}{Theorem}
\else
\fi

\ifarxiv
\else
\AtBeginDocument{%
  \providecommand\BibTeX{{%
    \normalfont B\kern-0.5em{\scshape i\kern-0.25em b}\kern-0.8em\TeX}}}
\fi

\begin{document}

\ifarxiv
\title{Diversified Ensembling: An Experiment in \\ Crowdsourced Machine Learning}
\author[1]{Ira Globus-Harris\thanks{Work done during an internship at Amazon}}
\author[1]{Declan Harrison\thanks{Work done during an internship at Amazon}}
\author[2]{Michael Kearns}
\author[2]{Pietro Perona}
\author[2]{Aaron Roth}
\affil[1]{University of Pennsylvania}
\affil[2]{AWS AI Labs}
\fi

\maketitle

\begin{abstract}
  Crowdsourced machine learning on competition platforms such as Kaggle is a popular and often effective method for generating accurate models. Typically, teams vie for the most accurate model, as measured by overall error on a holdout set, and it is common towards the end of such competitions for teams at the top of the leaderboard to ensemble or average their models outside the platform mechanism to get the final, best global model. In \cite{globus2022algorithmic}, the authors developed an alternative crowdsourcing framework in the context of fair machine learning, in order to integrate community feedback into models when subgroup unfairness is present and identifiable. There, unlike in classical crowdsourced ML, participants deliberately {\em specialize} their efforts by working on subproblems, such as demographic subgroups in the service of fairness. Here, we take a broader perspective on this work: we note that within this framework, participants may both specialize in the service of fairness and simply to cater to their particular expertise (e.g., focusing on identifying bird species in an image classification task). Unlike traditional crowdsourcing, this allows for the diversification of participants' efforts and may provide a participation mechanism to a larger range of individuals (e.g. a machine learning novice who has insight into a specific fairness concern). We present the first medium-scale experimental evaluation of this framework, with 46 participating teams attempting to generate models to predict income from American Community Survey data. We provide an empirical analysis of teams' approaches, and discuss the novel system architecture we developed. From here, we give concrete guidance for how best to deploy such a framework.
\end{abstract}

\section{Introduction}
Competition platforms are a popular framework for generating accurate machine learning models through communal efforts. Kaggle is the most popular of these ``crowdsourced" machine learning platforms, boasting fifteen million user accounts and thousands of competitions to date.\footnote{\url{https://www.kaggle.com/}} Companies and non-profits use the platform to publicly host competitions for learning tasks, often with rewards for the team with the highest performing model. One benefit of crowdsourcing models is that it gives a wider community access to the model development process: Kaggle, for instance, has been considered a mechanism for the ``democratization" of data science to a broader audience, particularly in the context of crowdsourced models for tasks with societal utility \cite{chou2014democratizing}. However, due to the standard structure of these competition frameworks, they do not truly leverage the expertise of all the competitors, and fail to explicitly align improvements in model fairness with competition success. Here, we implement and provide an empirical analysis of an alternate framework which provides such mechanisms.

In \cite{globus2022algorithmic}, the authors provide an alternative algorithmic framework for crowdsourcing machine learning models, which we implement here. Their framework was specifically designed for contexts where unfairness, in the form of disparate accuracy of models across identifiable subgroups of the distribution, is of concern, and where a model would be considered ``fair" if the model's error on each group is close to the Bayes optimal error on that group.\footnote{Note that this is \textit{not} a constrained optimization style notion of fairness, where, e.g., false positives are constrained to be equal across groups, and instead corresponds to group minimax fairness \citep{martinez2020minimax,diana2021minimax}. While in some cases a constraint-based approach to fairness may be more appropriate than a minimax style approach, we note that in many contexts, one might argue that when the Bayes optimal model is substantially different across groups, (un)fairness is now a question of data engineering and perhaps different features and/or more data should be collected to mitigate performance differences across the subgroups of interest.} In this framework, competitors compete against a global model $f$. At each round, they submit a function defining a group $g$ and a model $h$ which they claim has improved error compared to $f$ when restricted to the group $g$ (although $h$ need not improve on $f$ overall). If it does, as verified on a holdout set, then this pair $g$ and $h$ are incorporated via a natural ensembling technique into the model $f$, and this updated $f$ is used in subsequent rounds of the competition. The framework has attractive theoretical guarantees: each update decreases the overall error of the model, and so it is guaranteed to quickly converge to a state such that \emph{either} $f$ is Bayes optimal, or else no competitor can distinguish it from Bayes optimal using any $(g,h)$ pair. Moreover, the updates can be made in a way so that error is monotonically decreasing not just overall, but simultaneously on all of the groups $g$ identified in the competition so far\footnote{For formal statements, see Theorems 10, 12, and 14 in \cite{globus2022algorithmic}.}. Note that this approach is more general than the standard ``Kaggle'' design, which corresponds to a competition where teams always submit $(g,h)$-pairs where $g(x) \equiv 1$. 

This approach addresses two ways in which Kaggle-style crowdsourcing platforms fail to optimally direct the participants' efforts. In standard competitions, the final model is reflective of individual teams' efforts rather than a communal goal: one team to unilaterally ``wins'' the competition by proposing the most accurate model. It may well be that there are subregions of the dataset on which the best model is \emph{not} the winning model, but another competitors'. Thus, Kaggle does not truly leveraging each competitors strengths, nor does it provide a mechanism for competitors to specialize on specific subtasks. This is particularly a failure from the perspective of democratizing data science, as we wish to provide mechanisms for individuals who have specific real-world insights or expertise relevant to the particular machine learning task to contribute to model development. Ensembling the different competitors' models into one final model is a natural partial solution to this shortcoming of standard crowdsourcing, as it allows for specialization in model development. In practice, winning teams in crowdsourcing competitions often do suchmodel ensembling internally in an ad hoc way in order to leverage the strengths of different models: anecdotally, of the last eight Kaggle competitions with monetary prizes where models were published post-competition, five explicitly used some form of model ensembling (\cite{asl-fingerspelling, icr-identify-age-related-conditions, google-research-identify-contrails-reduce-global-warming, hubmap-hacking-the-human-vasculature, 2023-kaggle-ai-report, predict-student-performance-from-game-play, benetech-making-graphs-accessible, godaddy-microbusiness-density-forecasting, vesuvius-challenge-ink-detection, image-matching-challenge-2023}). The framework of \cite{globus2022algorithmic} explicitly builds this ensembling into the competition format, rather than relying on ad hoc ensembling outside of the competition framework.

A second motivation is that this ensembling method provides a mechanism to identify issues of unfairness and bias, and allows competitors to specialize. In the Kaggle-style framework, the only objective of interest is overall model accuracy. This reduces participants' incentives to focus on small regions of the distribution where the model performance is sub-optimal: efforts explicitly identifying and correcting bias on small groups only pay out if your model wins the competition. Companies need reporting (and reward) mechanisms for when individuals identify cases where their models underperform on groups of interest. In the framework of \cite{globus2022algorithmic}, such improvements are rewarded, and the goal of overall accuracy is explicitly aligned with optimal performance on subgroups.

\textbf{Results} We provide an empirical case study of the general framework for crowdsourced machine learning proposed by \cite{globus2022algorithmic}. In a real competition between 46 teams consisting of students at a major American university, we find that the final model outperforms all competitors' models individually (which is only possible because of the explicit ensembling in the competition design), and that competitors leveraged the ability to specialize. This specialization was done through a combination of algorithmic and manual data engineering approaches, and most teams attempted to use contextual knowledge of the task (measuring income) in order to make improvements to their models. We describe the novel platform infrastructure we implemented in order to host the competition and discuss the nuances of constructing such an architecture in a scalable manner. We discuss practical challenges and lessons learned from hosting such a system, from denial of service attacks to setting acceptance criteria in order to properly incentivizing later engagement in the competition, as well as usability challenges. 

\textbf{Related Work} The original model ensembling method in \cite{globus2022algorithmic} was designed as a ``bias bounties" competition, and was framed specifically as a method to combat unfairness. Here, we consider their framework more generally as a method to do crowdsourced machine learning while leveraging the strengths of all teams' models, rather than purely a method to combat unfairness. While the work of \cite{globus2022algorithmic} contained some preliminary experimental results in a very simplified framework, they did not include any true cross-team experiments in which a global model is built using multiple teams' contributions. Here, we provide the software architecture necessary to run such a crowdsourcing competition.

The ``bias bounty" idea used in \cite{globus2022algorithmic} dates at least to a 2018 Vice editorial \citep{bountyarticle}, and versions of the idea have been put into practice. In 2021, X (previously known as Twitter) released a bias bounties competition on their image cropping algorithm at DEFCON \citep{twitterbounty}. This competition had a substantially different framework than what is suggested here: a small number of competitors submitted written proposals, which were judged by a panel. Here, we consider larger-scale competitions where such empirical judging may be intractable (or legally fraught). Additionally, both the algorithmic framework of \cite{globus2022algorithmic} and the system design discussed here tackle adversarial behavior by competitors in crowdsourcing competitions. Other work along these lines in more traditional crowdsourced machine learning include work on exploiting data leakage (and prevention mechanisms) in crowdsourcing by \cite{narayanan2011} and \cite{kaufman2012leakage}, and mechanisms for preventing overfitting in leaderboard-based competitions by \cite{moritz-ladder}, among others. Finally, the algorithmic framework proposed by \cite{globus2022algorithmic} itself was independently and concurrently developed by \cite{tosh2022simple}. 

Another similar idea to bias bounties is the ``red teaming" of generative models to identify failure modes (e.g. poor performance on subgroups, leakage of confidential information, or code injection), which has recently been popularized: notably DEFCON 31 had a large redteaming event \citep{defcon}. The goal of this form of redteaming is to identify model vulnerabilities, but does not aim to fix them. In comparison, our approach offers a principled way to incorporate fixes into a model, with formal optimality guarantees about the resulting ameliorated model.

\textbf{Limitations} The primary limitation of the crowdsourcing framework itself is that it gives no guidance for how to find improvements to the model. This is expected of any crowdsourcing framework: the competitors' goal is to identify improvements, and as the competition progresses, finding these improvements becomes more challenging. We note that while competitors may specialize and focus on such subgroups, they could instead consider improvements on the entire model. As such, this framework is only more general than the standard crowdsourcing framework: when useful, competitors may specialize, but otherwise can focus on model-wide improvements. 

We emphasize that our framework is a proof-of-concept and that the recommendations we draw for future competitions are empirical in nature: we run a single competition, so are unable to make any statistical conclusions. Our competitors were given a relatively straightforward learning task on a tabular dataset where subgroups could be easily identified; in more complex tasks such as image identification, identifying subgroups may be more challenging. 

Competitors were also students, not machine learning experts, and sometimes made seemingly counter-intuitive choices in their approaches, which we discuss in greater detail in Appendix \ref{ap:emp}. We hope these observations may be more generally useful: one primary challenge of crowdsourcing as a means for democratizing data science is in the accessibility and usability of tools to those without computer science expertise \cite{chou2014democratizing}. We hope our observations may be used in subsequent crowdsourcing competitions and platforms in order to improve access and to avoid usability pitfalls. 

\section{Preliminaries and Background}\label{preliminaries}
For the purposes of our empirical study, we consider a competition for a regression problem. Formally, we consider a prediction task over a distribution of labelled examples $\Z = \X \times \Y$, where $\X$ are features and $\Y \in \mathbb{R}$ are real-valued labels that will be predicted by a model $f$. Let $\D \in \Delta Z$ be the joint distribution over features and labels, and let $D \sim \D^n$ denote a finite set of $n$ labeled samples drawn i.i.d.\ from the distribution $\D$. We measure performance of the model $f$ by its mean squared error over the distribution, where the error for a single prediction $f(x)$ is defined as $\ell: \X \times Y \rightarrow \mathbb{R}; \ \ell(f(x), y)) = (f(x) - y)^2$, and average loss is denoted $\Loss(\D, f) = \mean_{(x,y) \sim \D} [\ell(f(x), y)]$. 

In \cite{globus2022algorithmic}, the authors propose an algorithmic framework termed a ``bias bounty" for iteratively patching a predictor $f$ when regions with provably suboptimal performance are identified by auditors (See also \cite{tosh2022simple} for a very similar proposed algorithm that they call ``Prepend''). Formally, auditors are tasked with constructing $(g,h)$-pairs of functions: a \textit{group} indicator function $g : \mathcal{X} \to \{0,1\}$ and a \textit{hypothesis} predictor $h : \mathcal{X} \to \mathcal{Y}$. We define the group loss of a predictor $f$ on $g$ as $\Loss(\D, f, g) = \mean_{(x,y) \sim D}[(f(x)-y)^2 \vert g(x)=1]$. We let $w = \mean_{\D}[g(x)]$ be the \textit{weight} of group $g$ over $\D$. If a competitor is able to construct a $(g,h)$-pair such that $w (\Loss(\D, f,  g) - \Loss(\D, h, g)) > \alpha$, then the pair is \textit{accepted} and the model $f$ is \textit{updated}.

\begin{figure}[H]
\begin{centering}
\includegraphics[scale=0.9]{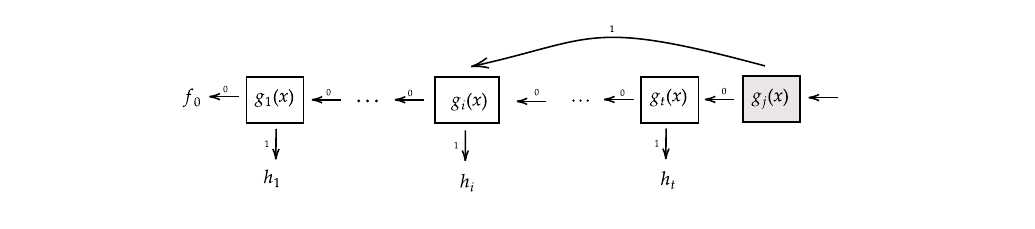}
\end{centering}
\caption{Model ensembling procedure for submitted models. In gray, a repair node which is created when the $(g_t, h_t)$ update increases error on group $g_j$. The best previous version of model for $g_j$ has been tracked, and the model is repaired to point to this.}
\label{fig:pdl}
\end{figure}

The algorithm for ensembling the model $f$ with the new model $h$ constructs a type of decision list with base node $f_0$, as shown in Figure \ref{fig:pdl}: when a $(g,h)$ pair is accepted, a new decision node is prepended to the structure with group inclusion as the test on the prepended node. Then, the updated model $f'$ will predict $h(x)$ for all instances such that $g(x) = 1$, and $f(x)$ otherwise. We can iterate this process, allowing competitors to search for new $(g,h)$-pairs to reduce  error on $f'$. Let ($g_{i}, h_{i}$) denote the $i^{th}$ accepted pair to the model, and let $f_{i}$ denote the model after it has prepended $(g_{i}, h_{i})$.

Since the proposed groups may not be disjoint, it is possible than an update $(g_{k}, h_{k})$ may improve performance on group $g_{k}$ while decreasing performance on an earlier group $g_{i}$ due to group intersections. In order to avoid this, the model is iteratively patched after updates: for each $g_{i}$ with $i \leq j < k$, the model $f_{j}$ which performs best on $g_{i}$ tracked. After each update, if the model's error on any previously identified group has gotten worse, then the model is \textit{repaired} by prepending a node with test for group inclusion of $g_{i}$ which ``points" to $f_{j}$ if $g_{i} = 1$ and to $f_{k}$ otherwise. In Figure \ref{fig:pdl}, the gray node is one such repair. This implies that once a $(g,h)$ pair is accepted into the overall model, the error rate on $g$ is non-increasing. We denote this as the ``repair" procedure, and can make the following formal guarantee:

\begin{theorem} [\cite{globus2022algorithmic}] 
\label{thm:bb}
Let $w_i = \mean[g_i(x)]$ and let $\Delta_i = \Loss(\D, h_i, g_i) - \Loss(\D, f_i,  g_i)$.  If all accepted $(g_i, h_i)$ satisfy $w_i \Delta_i > \alpha$, then at most $1/\alpha$ submissions may be accepted, including repairs. Furthermore, if a group is introduced at round $i, \{\Loss(\D, h_i, g_i) - \Loss(\D, f_j,  g_i); j > i \}$ is monotononically decreasing. 
\end{theorem}

In practice, we cannot verify improvements or track the group losses necessary for repairs over distributional loss. Instead, a sample of validation data, not accessible by competitors, is used to calculate all losses, and the above Theorem \ref{thm:bb} is generalized for in-sample use.\footnote{See for example Theorems 12 and 14 in \citep{globus2022algorithmic}}.

\section{Empirical Study}
\label{empirical-study}
We describe an empirical study conducted in an elective computer science course with a focus on algorithmic fairness at a prominent American research university during the spring semester of 2023. The study was deemed IRB exempt, and all students whose work is included in the subsequent analysis signed the consent form in Appendix \ref{ap:consent}. The assignment instructions are given in Appendix \ref{ap:proj-desc}. Students were not monetarily rewarded, and the assignments were graded without knowledge of who had agreed to have their work included in the study. Identifying markers for participants have been removed.

\subsection{Data Set and Prediction Task}
Competitors worked on a regression task predicting annual income for individuals in Southern US states earning between \$0 and \$100,000, based on ACS PUMS data derived from the Python Folktables \cite{ding2021retiring} package. The dataset contains 485,906 instances with twenty-one features, including sensitive attributes such as sex, race, age, and disability status. The full list of features included, along with their ACS encoding, are listed in Appendix \ref{ap:data-features}. Training, validation, and test splits were created with  70\%, 15\%, and 15\% weights respectively. Training data was distributed to students, validation data was used to determine acceptance for model updates, and test data was used for post hoc model evaluation.

\subsection{Competition Framework}
The study contained a total of one hundred thirty-nine graduate and undergraduate students from various STEM disciplines, predominantly computer science and data science. Participants were split into forty-five teams of three to four students for the project, which students had slightly over a month to complete. The competition was structured to have teams compete in two ways.

The first way teams competed exactly mirrors the crowdsourcing framework described above: teams all worked to update a central, constantly up-to-date model, which we will call the \textit{global} model. This model initially began as a relatively low-error gradient-boosting regressor trained by the course staff. Teams were given the initial training predictions from the global model, and tasked with constructing $(g,h)$-pairs to improve (subgroup) accuracy. If a team's $(g,h)$-pair was accepted, they were rewarded with points proportional to the reduction in the model's overall validation error. The model was updated to incorporate the $(g,h)$-pair, and a notification was sent to all participants containing the reduction in error as well as the global models' updated training predictions.

Every time a team submitted a $(g,h)$ pair to this global competition, that pair was also submitted to a \textit{local} version of the competition specific to that particular group, which built an ensembled model using only that single teams' submissions. The initial local model for each team was a depth one decision tree fit on the training data, essentially predicting the mean label for all instances. Teams' local models were evaluated based on the validation error rate, and a leaderboard was displayed throughout the competition for teams to view the relative performance of their local model against others. 

Of the 6914 $(g,h)$-pairs that the forty-five teams submitted, 3137 of them were submitted by Team 7, who automated a brute force approach. We discuss their approach briefly in \ref{sec:security}, but omit their submissions in our analysis in order to give a clearer picture of the other teams' efforts.

\subsection{Analysis of the Competition}
\label{sec:analysis}

\begin{figure}[H]
\begin{center}
\includegraphics[width = .96\linewidth, trim  ={1cm 1cm 1cm 1cm}, clip]{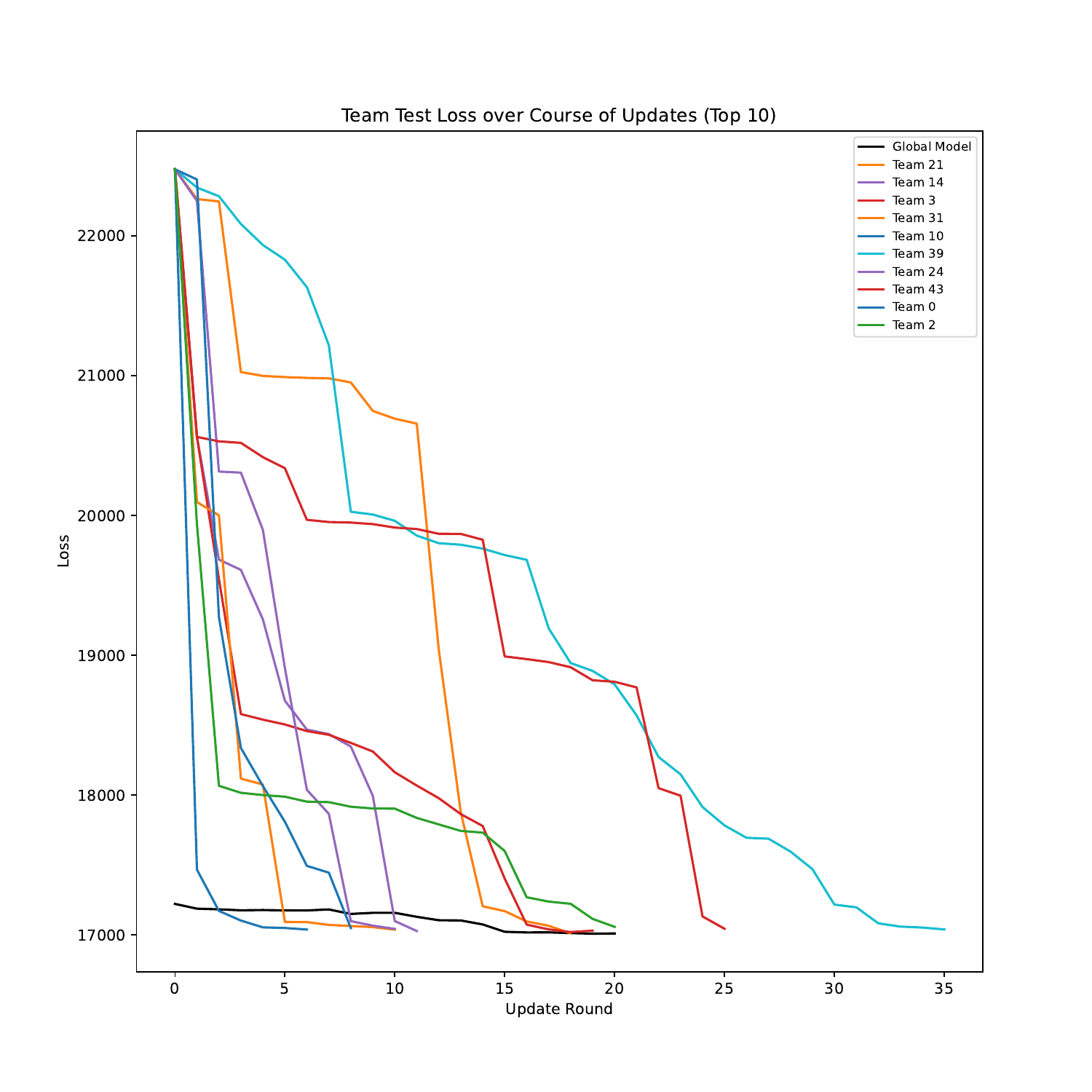}
\caption{Tracking test error of the top ten teams' local models per accepted update vs. global model. In the legend, the teams are ordered by the performance of their local model top-to-bottom; see Table \ref{tab:results-table} for final rankings of teams' models.}
\label{fig:team-ranking}
\end{center}
\end{figure}

\begin{figure}[H]
    \includegraphics[width=\linewidth]{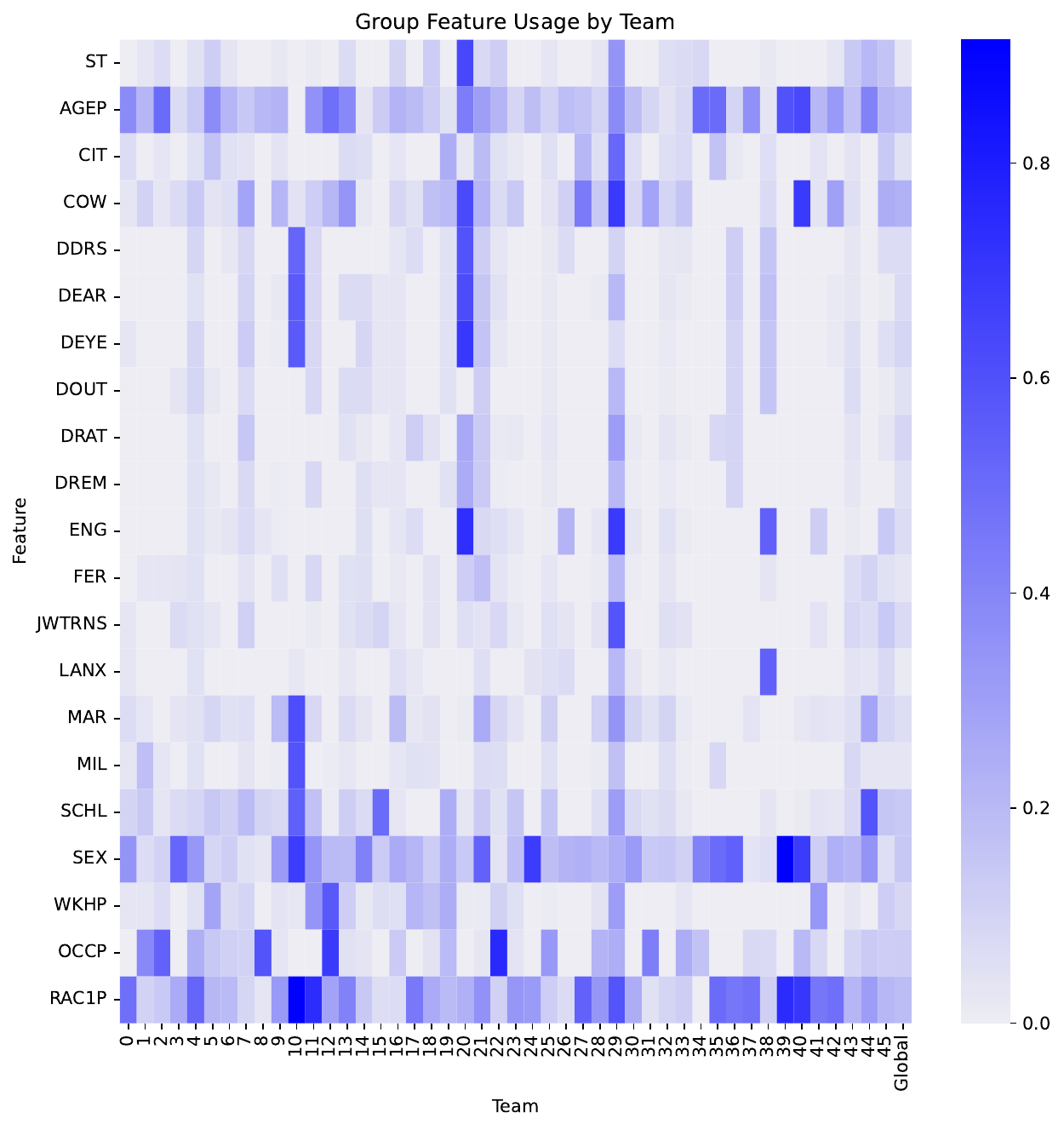}
    \caption{Frequency of feature usage in group submissions by team. The features listed on the $y$-axis are explained in more depth in Appendix \ref{ap:data-features}.}
    \label{fig:feature-usage-by-team}
\end{figure}

We provide empirical observations about the approaches employed by students in the competition. Based on these observations, we conclude with a concrete set of suggestions for how to create competition designs which are maximally usable for competitors.

\textbf{The global model outperformed all local models:} Ensembling the crowdsourced models produced a better model than individual teams' efforts. Eleven teams (including Team 7) managed to submit a total of twenty accepted updates to the global model, which outperformed all teams' local models at the completion of the study. We provide the tracked train/validation/test errors per update for the global and local models as well as a numerical table of final errors in Appendix \ref{appendix-graphs}. Specifically, the best team had an overall squared error of 17012.32 on the holdout test data, while the global model had squared error of 17010.49--a slight improvement but an improvement nonetheless. This is possible only because the global model is explicitly ensembling contributions from multiple teams.

\textbf{Specializing models to subgroups helped:} Teams could choose to compete using only traditional Kaggle-style updates where the submitted $g$ represents the entire dataset. One team (team 10) did so: each of their six successful updates fully replaced their previous models. Their final model performed well, 5th out of all groups, as shown in Figure \ref{fig:team-ranking}. Thus, treating the competition as purely a Kaggle-style exercise was a relatively competitive strategy, but was not \textit{the} most competitive strategy among our participants. This is further demonstrated in Figure \ref{fig:acc-weight} in Appendix \ref{ap:extra-plots}; which shows that accepted updates usually somewhat specialized. Furthermore, while the global model did accept some later updates over the entire dataset, theyeach triggered the repairs described in Section \ref{preliminaries}. In other words, there were models submitted earlier in the competition that performed better on their associated subgroup than later updates, and thus replacing the entire model with a later one naively would have caused an increase in error on those groups --- even as it led to decreased overall error. The automated repair method corrects for this, which involves \emph{re-ensembling} the submitted model with previous submissions, even when the most recently accepted submission corresponds to a trivial ``group'' corresponding to the entire dataset.  Thus \emph{even when competitors who choose to ignore the ability to target subgoups $g$ and instead submit Kaggle-style updates have their updates accepted, the resulting model is still an ensemble, which is accuracy-improving}.

\begin{figure}[H]
    \begin{center}
    \includegraphics[scale=0.3]{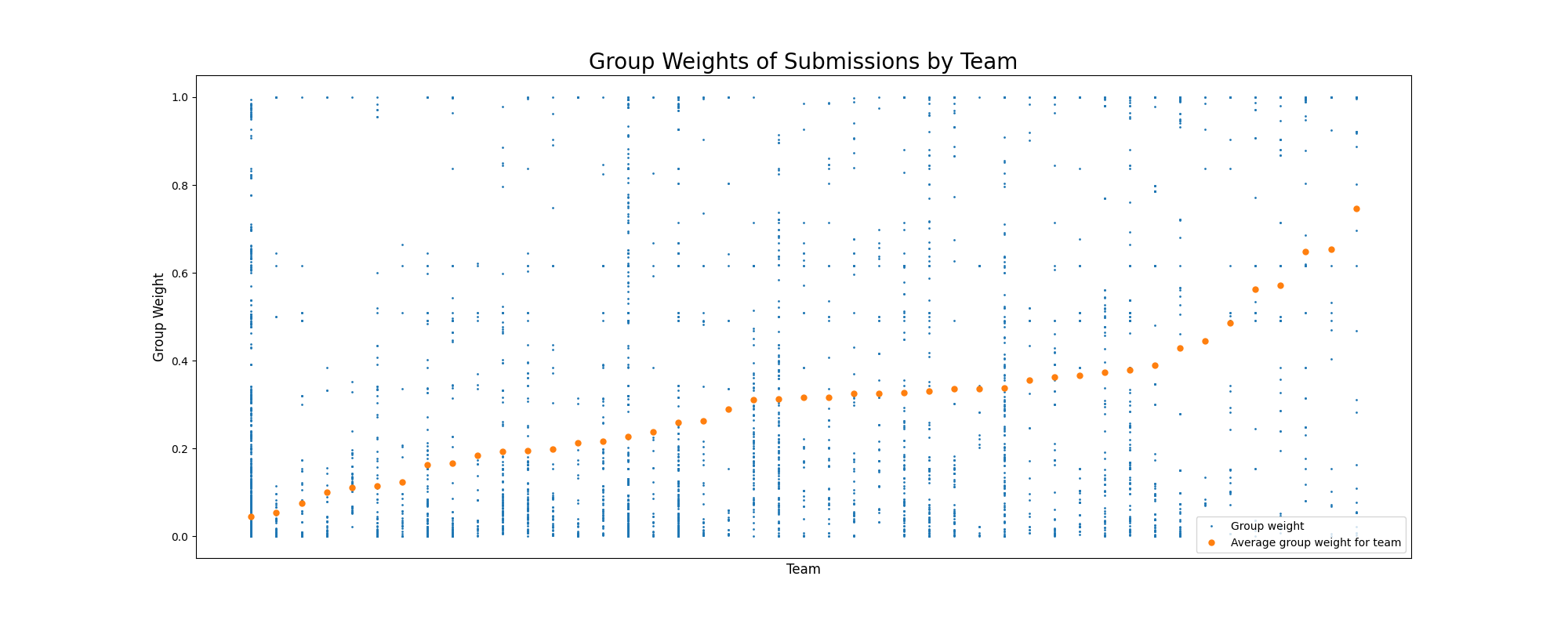}
    \caption{Distribution of weights ($\mean_{\D}[g(x)]$) of groups submitted by teams, with teams sorted by average group weight. Each vertical line corresponds to a single team, and the blue dots are the weights of the groups they submitted. The orange dots are the average weight of groups submitted by the teams.}
    \label{fig:weights-by-teams}
    \end{center}
\end{figure}

\textbf{Most teams specialized, and did so differently:} Over the course of the competition, a total of 896 updates were made across the 46 teams, ranging from 5 to 42 updates per team. As seen in Figure \ref{fig:weights-by-teams}, the sizes of groups submitted on average differed significantly between teams, indicating that teams broke the problem up differently. More explicitly, in Figure \ref{fig:feature-usage-by-team}, we see the distribution over submissions of which features teams used to define their groups $g$, measured by frequency. A complete mapping of the feature acronyms to their meanings is provided in Appendix \ref{ap:data-features}; in the figure while features such as race (RAC1P), binary sex (SEX), and age (AGEP) were commonly targeted by a majority of teams, we see subsets of teams focusing on other features such as education level (SCHL) and disability status (primarily DDRS, DEAR and DEYE).

\textbf{Most teams employed a combination of manual, automated, and learned approaches:} Teams had multiple approaches for identifying groups of datapoints to make improvements on. They could \textit{manually} identify regions where the model performed poorly, e.g. by conditioning on different features to find regions the model might be specified to perform poorly on. Or, they could use an \textit{algorithmic} approach, e.g. specifying some class $\cG$ of possible groups and then trying to learn the group $g$ in $\cG$ where the current model performed worst. Teams were given a file with potential algorithmic approaches they could attempt and were encouraged to explore their own methods.\footnote{Algorithmic methods file is listed in code repository listed in Appendix.} Purely algorithmic approaches such as learning clusters of datapoints where performance was suboptimal were largely less successful: only three of the 19 $(g,h)$-pairs accepted to the global model were automated, and in general automated updates had an acceptance rate of 13.6 percent, as opposed to the 25.8 percent acceptance rate for manual updates. Automated approaches were also less popular in terms of overall number of submissions: only 21.8 percent of the submitted updates were automated approaches from 37 unique teams. 

\textbf{Groups were often chosen using contextual knowledge:} One approach many (roughly 85 percent) of the teams had was to condition across the feature space using their knowledge of the context of the prediction task. For instance, many teams chose to examine subgroups related to gender or race. In particular, of the 821 submissions that used race as a predicate for the group, 327 (39\%) subsetted over African Americans, and when binarized sex was a predicate, it was overwhelmingly (96\% of the time) used to subset for female-identified individuals. In their write-ups, students stated that they believed that these subgroups might, due to systemic bias or discrimination, have disparate pay, and hence these features might be leveraged to improve the model. The efficacy of this approach is arguable, as discussed in greater detail in Appendix \ref{ap:emp}: specialization to narrow subgroups did help teams find updates, but often this required careful examination of the data.

\textbf{Finding Later Updates and Identifying Promising Subgroups is Hard}
 As in any competition, we expected teams to struggle to find improvements to the model as the competition wore on. This prediction was correct: at the end of the month only eleven of forty-five teams managed to have updates accepted on the global model. One primary challenge that students wrote about was how to balance specialization with generalization: if the subgroups of interest are too small, then restricting training to that subgroup will likely overfit. Specialization only helps if you have reason to believe that this subgroup has some generalizable structure that the larger model has yet to find, and knowing when this might occur is challenging. 

\textbf{Final Models Perform Similarly While Making Different Predictions} While the model desired from a large scale competition would be crowdsourced from all competitors, we constructed local models to reflect efforts from individual teams for purposes of grading. However, these local models produced an interesting phenomena; the leaderboard for the most accurate local models shows a narrow margin compared to the crowdsourced global model yet these models make substantially different predictions. In Figure \ref{fig:top-5-diff}, we plot the absolute difference in predictions between the global model and the top five local models to see a non-trivial density of instances for which the models disagree.

\section{Platform Design}\label{Platform Design}
In this section, we introduce a platform for hosting the competitions. The traditional method for authenticated user interaction with a server is to develop a \textit{full stack} solution; a comprehensive system covering web development, database management, back end software, and more. Naturally, these systems require expertise to build and have serious implications when incorrectly constructed. Context management services, such as \textit{Wordpress}, reduce these requirements but have monthly fees and frequently contain vulnerabilities.\footnote{\url{https://cve.mitre.org/cgi-bin/cvekey.cgi?keyword=wordpress}} Instead, our platform leverages GitHub to host competitions, gaining the security and web interface inherent to GitHub. The open-source package and detailed installation instruction are available for download at \url{https://anonymous.4open.science/r/Diversified-Ensembling-Template-FF3E/README.md}.

\subsection{Platform Design}
In our construction, competitions are hosted on private GitHub repository cloned by the organizer from our package template. During the package installation, the organizer connects a backend computation source, e.g. an AWS EC2 instance or personal Linux machine, to the repository using a continuous integration and deployment tool, \textit{GitHub Actions}. By utilizing \textit{Actions}, we ``solve" two major components of a secure web-server stack: data base management and a front-end interface.

Since competitions are run inside \textit{private} repositories, an organizer may easily add and remove participants through the GitHub GUI, avoiding the expertise needed to manage an SQL database. Additionally, the competition platform gains GitHub's user authentication protocols; accessing a competition implies repository read permissions and there are currently no known vulnerabilities for reading private repositories without appropriate permissions. 

\begin{figure}
\begin{center}
\includegraphics[width = .9\linewidth]{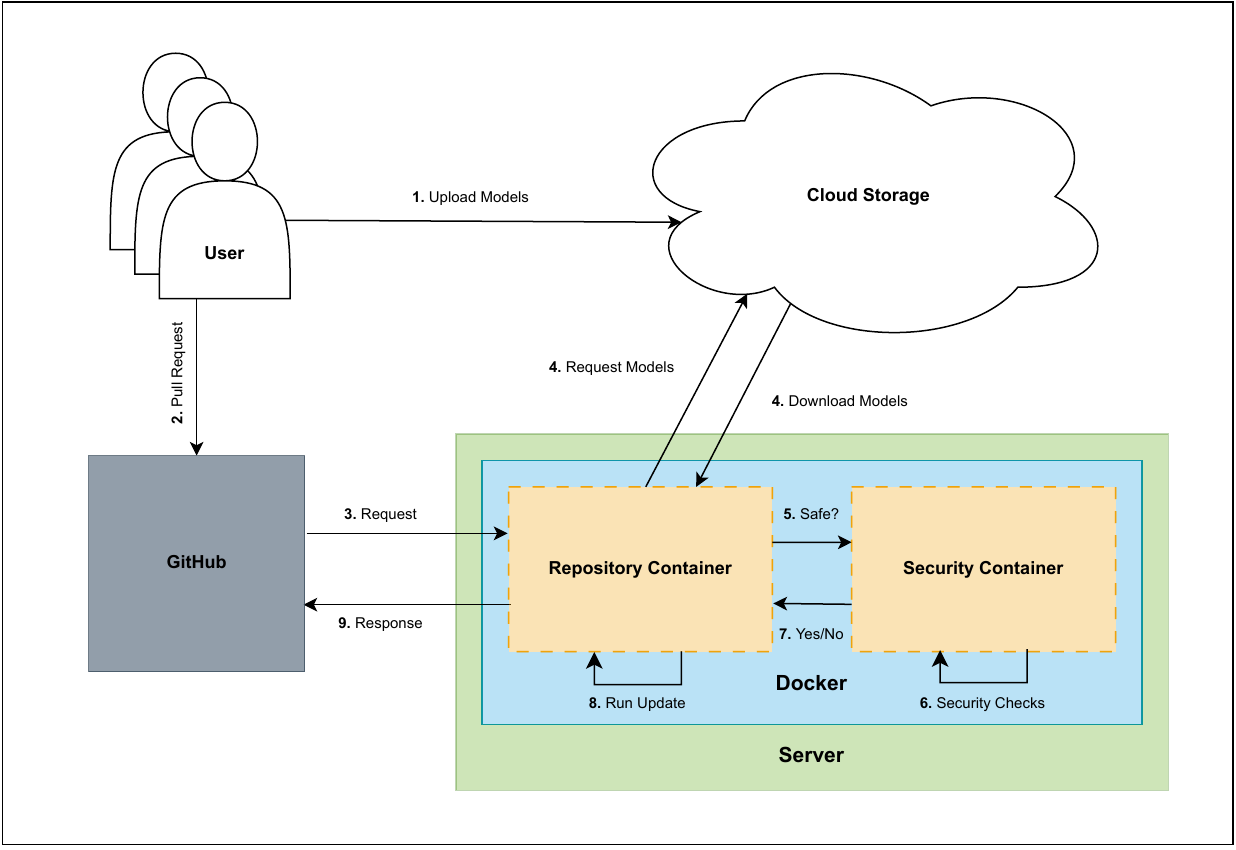}
\caption{Protocol for $(g,h)$ pair submissions from participants to a competition repository.}
\label{fig:infrastructure}
\end{center}
\end{figure}

Due to the expected background of participants in ML competitions, we assume basic familiarity and likely comfort with GitHub. One challenge of a competition is the global model needs to be  accessible to competitors, be constantly updated, and be able to accept competitors $(g,h)$-pair submissions. GitHub trivially solves these problems through \textit{push} and \textit{pull} requests to a repository. Our platform stores necessary competition information such as the current global model's training predictions and the leaderboard in such a repo, and participants interact with the repository to submit $(g,h)$ pairs through pull requests. The detailed format of these pull requests is described in Appendix \ref{ap:deep-dive}, but, at its core, participants upload models to a cloud provider (like Google Drive or AWS S3) and provide links to each models in their request. Feedback is provided to participants via comments on pull requests and any successful update forces a push to the repository with updated model information. The submission-feedback loop with backend protocol is shown in Figure \ref{fig:infrastructure}.

As an added bonus factor, GitHub allows users to create websites from HTML and Markdown code in their repository, which allows us to easily integrate a public leaderboard and general competition information in a digestible format for users. 

\subsection{Security Protections \& Vulnerability Concerns}
\label{sec:security}
In order to evaluate whether or not a submitted $(g,h)$-pair ought to be integrated into the model, a competitors' code has to be run on the server on the validation data. As a result, there is a risk of an adversarial competitor's malicious code being run on the host machine. While this risk cannot be entirely mitigated, we describe baseline security measure implementations and describe future methods for securing systems. 

First, GitHub is used to authenticate all teams. Thus, the primary layer of security is that the competitors are verified. Within a classroom context this worked well: if a student had submitted malicious code, we would have been able to trace it back to the student and their grade would have been impacted, which was a suitable disincentive. In deployment outside this setting, we suggest hosts require users to sign a legal agreement prior to participation as is standard practice in ML competitions. 

Secondly, the competition is run within a user-mode Docker container with limited kernel privileges. When a (g,h) pair is submitted from a participant, the files are downloaded into the Docker container. The models are passed to a second Docker container with no internet access to the run the security checks from the following paragraph to determine if malicious code is identified. This protection ensures the host machine is not affected from code run in the competition but does not protect the repository files.   

Lastly, models are checked on load for malicious behavior to include loading or importing unnecessary packages e.g. \textit{sys} or \textit{os}. Since models are saved and loaded as serialized byte streams, evaluating model intentions is a difficult task as it is done at the opcode level via a disassembler. We intend to increase the breadth of checks done in this section through updates to our package. 

In practice, the primary security issue we had was an inadvertent denial of service attack by the team who automated submissions. This led to a long queue for other teams' pairs to be verified. In the future, we would implement submission limits per team (per day and overall), to avoid this.

\section{Lessons Learned}
We provide a general overview of lessons learned from the framework's deployment, both from a systems perspective and for optimizing competitors' engagement with the platform.

\textbf{Distribute Environment Files.} A major factor in this competition is being able to pass ML models between participants and the server. Distribution of a makefile that constructs a virtual environment or a Docker container for participants to work helps this process run smoothly. While this is a relatively standard practice for running code across platforms, we emphasize its' importance: using different versions of packages (or Python) caused major frustrations for participants as it often resulted in denial of submissions for ``security'' errors.

\textbf{Limit daily submissions.} As discussed in greater detail in Section \ref{sec:security}, Team 7 took an automated, brute-force approach that caused a Denial of Service. In order to prevent this, as well as to incentivize competitors to only send in submissions that they truly believe are competitive, it would be wise to limit the number of daily submissions. Additionally, this reduces overfitting to the validation set, as it prevents hillclimbing.

\textbf{Prime competitors to think critically about group identification.} Since competitors were students in a class which focused largely on fairness in machine learning, they were primed to think about the effect of societal factors. As discussed in Section \ref{sec:analysis}, this led students to, e.g., almost exclusively condition on binarized sex being female when considering sex as a predicate. While competitors should be encouraged to use their general knowledge to identify regions of disparate performance, doing so effectively may require careful data engineering, and looking for general performance improvements is also useful. 

\textbf{Setting the $\alpha$ threshold for acceptance is a tricky, data-dependent task.} The competition designers set a threshold $\alpha$ which determines how much of an improvement an update must incur in order to be accepted. While we do not have the counterfactual of setting different values for our $\alpha$, re-simulating the competition with higher or lower values led to lower performance and overfitting respectively. Before beginning a competition, hosts should try various threshold values for their given task and loss function with trusted users to observe what value gives the best generalization/performance balance.   

\textbf{Alter reward system to scale over time.} Competitors received points based on the amount of validation error decreased by an accepted update on the global model. Intuitively, teams that participated earlier in the study were able to make more updates to the global model: notably, Team 12 made the first seven updates to global model. However, as the competition progressed, the problem became more difficult and teams weren't as able to decrease the error of the model. In order to bolster effort throughout the duration of the competition, rewards could scale by both the amount of error reduced as well as the number of previous updates made or time since start of the competition.

\section{Dataset}
In the process of running the competition, we generated a database of nearly seven thousand $(g,h)$-pairs which competitors submitted, which may be of independent interest. The dataset consists of 6914 $(g,h)$-pairs of models which make predictions over the income task described in Section \ref{empirical-study}, where the $g$'s describe subsets of the distribution and the $h$'s are a variety of different models trained over those subsets. 

In general, large datasets of machine learning models may be of academic interest. Kaggle itself has a Meta Kaggle\footnote{\url{https://www.kaggle.com/datasets/kaggle/meta-kaggle}} dataset, which is a public dataset of competitions and submissions on the platform and which has been widely used, e.g. in \cite{kowald2019using} and \cite{roelofs2019meta}. We believe that, in a similar spirit, our dataset is also of use. In particular, the fact that it provides a large collection of subgroups in addition to models offers multiple use cases.

First, the fairness literature often assumes fairness guarantees are desired with respect to some rich class of groups $\mathcal{G}$. In practice, these papers usually either contain no experimental results at all, or results with respect to extremely minimal and usually disjoint groups: e.g. race, binarized sex, or the two-way marginals of race and binarized sex. Here, we have a much richer collection of many thousands of groups which span many features, and suggest that these may be used for benchmarking fairness approaches.

Secondly, we note that the ensembling framework proposed in \cite{globus2022algorithmic} was independently developed by \cite{tosh2022simple} in the context of multi-group learning. There, they provide an additional algorithm that frames multi-group learning as a form of sleeping experts, where each $(g,h)$-pair corresponds to an expert, and each expert is ``awake" when $g(x)=1$. This formulation is beneficial, as the reduction of sleeping experts to the offline setting gives an  algorithm leads to improved sample complexity compared to the decision-list style updates proposed in \cite{globus2022algorithmic}. However, this work is theoretical in nature, and does not provide experimental guarantees: in particular, it assumes that $\vert \cG \vert$ is finite and that $\ell \circ \cH$ has bounded pseudodimension, and requires computations over a large number of $(g,h)$-pairs. It is not clear how much improvement this methodology would give in practical settings. Our dataset offers one way of measuring this.

Thus, the dataset of model-pairs and its associated training and test data, may be used as both a form of fairness benchmark and as a mechanism to evaluate ensembling methods and expert learning more generally. 

\subsubsection*{Acknowledgements}

We give warm thanks to Peter Hallinan for many helpful conversations. 

\bibliographystyle{ACM-Reference-Format}
\bibliography{references}


\begin{thebibliography}{25}


\ifx \showCODEN    \undefined \def \showCODEN     #1{\unskip}     \fi
\ifx \showDOI      \undefined \def \showDOI       #1{#1}\fi
\ifx \showISBNx    \undefined \def \showISBNx     #1{\unskip}     \fi
\ifx \showISBNxiii \undefined \def \showISBNxiii  #1{\unskip}     \fi
\ifx \showISSN     \undefined \def \showISSN      #1{\unskip}     \fi
\ifx \showLCCN     \undefined \def \showLCCN      #1{\unskip}     \fi
\ifx \shownote     \undefined \def \shownote      #1{#1}          \fi
\ifx \showarticletitle \undefined \def \showarticletitle #1{#1}   \fi
\ifx \showURL      \undefined \def \showURL       {\relax}        \fi
\providecommand\bibfield[2]{#2}
\providecommand\bibinfo[2]{#2}
\providecommand\natexlab[1]{#1}
\providecommand\showeprint[2][]{arXiv:#2}

\bibitem[Andrews et~al\mbox{.}(2023)]%
        {benetech-making-graphs-accessible}
\bibfield{author}{\bibinfo{person}{Benji Andrews}, \bibinfo{person}{Hema Natarajan}, \bibinfo{person}{Maggie}, \bibinfo{person}{Ron Ellis}, {and} \bibinfo{person}{Ryan Holbrook}.} \bibinfo{year}{2023}\natexlab{}.
\newblock \bibinfo{title}{Benetech - Making Graphs Accessible}.
\newblock
\newblock
\urldef\tempurl%
\url{https://kaggle.com/competitions/benetech-making-graphs-accessible}
\showURL{%
\tempurl}


\bibitem[Blum and Hardt(2015)]%
        {moritz-ladder}
\bibfield{author}{\bibinfo{person}{Avrim Blum} {and} \bibinfo{person}{Moritz Hardt}.} \bibinfo{year}{2015}\natexlab{}.
\newblock \showarticletitle{The Ladder: A Reliable Leaderboard for Machine Learning Competitions}. In \bibinfo{booktitle}{\emph{Proceedings of the 32nd International Conference on Machine Learning}} \emph{(\bibinfo{series}{Proceedings of Machine Learning Research}, Vol.~\bibinfo{volume}{37})}, \bibfield{editor}{\bibinfo{person}{Francis Bach} {and} \bibinfo{person}{David Blei}} (Eds.). \bibinfo{publisher}{PMLR}, \bibinfo{address}{Lille, France}, \bibinfo{pages}{1006--1014}.
\newblock
\urldef\tempurl%
\url{https://proceedings.mlr.press/v37/blum15.html}
\showURL{%
\tempurl}


\bibitem[Carman et~al\mbox{.}(2023)]%
        {icr-identify-age-related-conditions}
\bibfield{author}{\bibinfo{person}{Aaron Carman}, \bibinfo{person}{Alexander Heifler}, \bibinfo{person}{Ashley Chow}, {and} \bibinfo{person}{Ryan Holbrook}.} \bibinfo{year}{2023}\natexlab{}.
\newblock \bibinfo{title}{ICR - Identifying Age-Related Conditions}.
\newblock
\newblock
\urldef\tempurl%
\url{https://kaggle.com/competitions/icr-identify-age-related-conditions}
\showURL{%
\tempurl}


\bibitem[Census(2022)]%
        {pums-dictionary}
\bibfield{author}{\bibinfo{person}{US Census}.} \bibinfo{year}{October 20, 2022}\natexlab{}.
\newblock \bibinfo{booktitle}{\emph{2021 ACS PUMS Data Dictionary}}.
\newblock \bibinfo{type}{Data Dictionary}. \bibinfo{institution}{US Census}.
\newblock
\urldef\tempurl%
\url{https://www2.census.gov/programs-surveys/acs/tech_docs/pums/data_dict/PUMS_Data_Dictionary_2021.pdf}
\showURL{%
\tempurl}


\bibitem[Chou et~al\mbox{.}(2014)]%
        {chou2014democratizing}
\bibfield{author}{\bibinfo{person}{Sophie Chou}, \bibinfo{person}{William Li}, {and} \bibinfo{person}{Ramesh Sridharan}.} \bibinfo{year}{2014}\natexlab{}.
\newblock \showarticletitle{Democratizing data science}. In \bibinfo{booktitle}{\emph{Proceedings of the KDD 2014 20th ACM SIGKDD International Conference on Knowledge Discovery and Data Mining, New York, NY, USA}}. \bibinfo{pages}{24--27}.
\newblock


\bibitem[Chow et~al\mbox{.}(2023a)]%
        {asl-fingerspelling}
\bibfield{author}{\bibinfo{person}{Ashley Chow}, \bibinfo{person}{Glenn Cameron}, \bibinfo{person}{Manfred Georg}, \bibinfo{person}{Mark Sherwood}, \bibinfo{person}{Phil Culliton}, \bibinfo{person}{Sam Sepah}, \bibinfo{person}{Sohier Dane}, {and} \bibinfo{person}{Thad Starner}.} \bibinfo{year}{2023}\natexlab{a}.
\newblock \bibinfo{title}{Google - American Sign Language Fingerspelling Recognition}.
\newblock
\newblock
\urldef\tempurl%
\url{https://kaggle.com/competitions/asl-fingerspelling}
\showURL{%
\tempurl}


\bibitem[Chow et~al\mbox{.}(2023b)]%
        {image-matching-challenge-2023}
\bibfield{author}{\bibinfo{person}{Ashley Chow}, \bibinfo{person}{Eduard Trulls}, \bibinfo{person}{Jevster}, \bibinfo{person}{Kwang~Moo Yi}, \bibinfo{person}{Sohier Dane}, \bibinfo{person}{Tanji Gou}, {and} \bibinfo{person}{Weiwei Sun}.} \bibinfo{year}{2023}\natexlab{b}.
\newblock \bibinfo{title}{Image Matching Challenge 2023}.
\newblock
\newblock
\urldef\tempurl%
\url{https://kaggle.com/competitions/image-matching-challenge-2023}
\showURL{%
\tempurl}


\bibitem[Chowdhury and Williams(2021)]%
        {twitterbounty}
\bibfield{author}{\bibinfo{person}{Rumman Chowdhury} {and} \bibinfo{person}{Jutta Williams}.} \bibinfo{year}{2021}\natexlab{}.
\newblock \bibinfo{title}{Introducing Twitter’s First Algorithmic Bias Bounty Challenge}.
\newblock
\newblock
\urldef\tempurl%
\url{https://blog.twitter.com/engineering/en_us/topics/insights/2021/algorithmic-bias-bounty-challenge}
\showURL{%
\tempurl}


\bibitem[Diana et~al\mbox{.}(2021)]%
        {diana2021minimax}
\bibfield{author}{\bibinfo{person}{Emily Diana}, \bibinfo{person}{Wesley Gill}, \bibinfo{person}{Michael Kearns}, \bibinfo{person}{Krishnaram Kenthapadi}, {and} \bibinfo{person}{Aaron Roth}.} \bibinfo{year}{2021}\natexlab{}.
\newblock \showarticletitle{Minimax group fairness: Algorithms and experiments}. In \bibinfo{booktitle}{\emph{Proceedings of the 2021 AAAI/ACM Conference on AI, Ethics, and Society}}. \bibinfo{pages}{66--76}.
\newblock


\bibitem[Ding et~al\mbox{.}(2021)]%
        {ding2021retiring}
\bibfield{author}{\bibinfo{person}{Frances Ding}, \bibinfo{person}{Moritz Hardt}, \bibinfo{person}{John Miller}, {and} \bibinfo{person}{Ludwig Schmidt}.} \bibinfo{year}{2021}\natexlab{}.
\newblock \showarticletitle{Retiring Adult: New Datasets for Fair Machine Learning}.
\newblock \bibinfo{journal}{\emph{Advances in Neural Information Processing Systems}}  \bibinfo{volume}{34} (\bibinfo{year}{2021}).
\newblock


\bibitem[Franklin et~al\mbox{.}(2023)]%
        {predict-student-performance-from-game-play}
\bibfield{author}{\bibinfo{person}{Alex Franklin}, \bibinfo{person}{David Gagnon}, \bibinfo{person}{Maggie}, \bibinfo{person}{Meg Benner}, \bibinfo{person}{Natalie Rambis}, \bibinfo{person}{Perpetual Baffour}, \bibinfo{person}{Phil Culliton}, \bibinfo{person}{Scott Crossley}, {and} \bibinfo{person}{Ulrich Boser}.} \bibinfo{year}{2023}\natexlab{}.
\newblock \bibinfo{title}{Predict Student Performance from Game Play}.
\newblock
\newblock
\urldef\tempurl%
\url{https://kaggle.com/competitions/predict-student-performance-from-game-play}
\showURL{%
\tempurl}


\bibitem[Globus-Harris et~al\mbox{.}(2022)]%
        {globus2022algorithmic}
\bibfield{author}{\bibinfo{person}{Ira Globus-Harris}, \bibinfo{person}{Michael Kearns}, {and} \bibinfo{person}{Aaron Roth}.} \bibinfo{year}{2022}\natexlab{}.
\newblock \showarticletitle{An algorithmic framework for bias bounties}. In \bibinfo{booktitle}{\emph{Proceedings of the 2022 ACM Conference on Fairness, Accountability, and Transparency}}. \bibinfo{pages}{1106--1124}.
\newblock


\bibitem[Howard et~al\mbox{.}(2022)]%
        {godaddy-microbusiness-density-forecasting}
\bibfield{author}{\bibinfo{person}{Addison Howard}, \bibinfo{person}{Archit Agarwal}, \bibinfo{person}{Ashley Chow}, \bibinfo{person}{Rantig}, \bibinfo{person}{Kellen~J Gracey}, \bibinfo{person}{Robert~JC Brown}, {and} \bibinfo{person}{Sohier Dane}.} \bibinfo{year}{2022}\natexlab{}.
\newblock \bibinfo{title}{GoDaddy - Microbusiness Density Forecasting}.
\newblock
\newblock
\urldef\tempurl%
\url{https://kaggle.com/competitions/godaddy-microbusiness-density-forecasting}
\showURL{%
\tempurl}


\bibitem[Howard et~al\mbox{.}(2023)]%
        {hubmap-hacking-the-human-vasculature}
\bibfield{author}{\bibinfo{person}{Addison Howard}, \bibinfo{person}{Jevster}, \bibinfo{person}{Katherine Gustilo}, \bibinfo{person}{Katy Borner}, \bibinfo{person}{Ryan Holbrook}, {and} \bibinfo{person}{Yashvardhan Jain}.} \bibinfo{year}{2023}\natexlab{}.
\newblock \bibinfo{title}{HuBMAP - Hacking the Human Vasculature}.
\newblock
\newblock
\urldef\tempurl%
\url{https://kaggle.com/competitions/hubmap-hacking-the-human-vasculature}
\showURL{%
\tempurl}


\bibitem[Kaufman et~al\mbox{.}(2012)]%
        {kaufman2012leakage}
\bibfield{author}{\bibinfo{person}{Shachar Kaufman}, \bibinfo{person}{Saharon Rosset}, \bibinfo{person}{Claudia Perlich}, {and} \bibinfo{person}{Ori Stitelman}.} \bibinfo{year}{2012}\natexlab{}.
\newblock \showarticletitle{Leakage in data mining: Formulation, detection, and avoidance}.
\newblock \bibinfo{journal}{\emph{ACM Transactions on Knowledge Discovery from Data (TKDD)}} \bibinfo{volume}{6}, \bibinfo{number}{4} (\bibinfo{year}{2012}), \bibinfo{pages}{1--21}.
\newblock


\bibitem[Kowald et~al\mbox{.}(2019)]%
        {kowald2019using}
\bibfield{author}{\bibinfo{person}{Dominik Kowald}, \bibinfo{person}{Matthias Traub}, \bibinfo{person}{Dieter Theiler}, \bibinfo{person}{Heimo Gursch}, \bibinfo{person}{Emanuel Lacic}, \bibinfo{person}{Stefanie Lindstaedt}, \bibinfo{person}{Roman Kern}, {and} \bibinfo{person}{Elisabeth Lex}.} \bibinfo{year}{2019}\natexlab{}.
\newblock \showarticletitle{Using the Open Meta Kaggle Dataset to Evaluate Tripartite Recommendations in Data Markets}.
\newblock \bibinfo{journal}{\emph{arXiv preprint arXiv:1908.04017}} (\bibinfo{year}{2019}).
\newblock


\bibitem[Lourenco et~al\mbox{.}(2023)]%
        {vesuvius-challenge-ink-detection}
\bibfield{author}{\bibinfo{person}{Alex Lourenco}, \bibinfo{person}{Brent Seales}, \bibinfo{person}{Christy Chapman}, \bibinfo{person}{Daniel Havir}, \bibinfo{person}{Ian~Janicki Janicki}, \bibinfo{person}{JP Posma}, \bibinfo{person}{Nat Friedman}, \bibinfo{person}{Ryan Holbrook}, \bibinfo{person}{Seth P.}, \bibinfo{person}{Stephen Parsons}, {and} \bibinfo{person}{Will Cukierski}.} \bibinfo{year}{2023}\natexlab{}.
\newblock \bibinfo{title}{Vesuvius Challenge - Ink Detection}.
\newblock
\newblock
\urldef\tempurl%
\url{https://kaggle.com/competitions/vesuvius-challenge-ink-detection}
\showURL{%
\tempurl}


\bibitem[Martinez et~al\mbox{.}(2020)]%
        {martinez2020minimax}
\bibfield{author}{\bibinfo{person}{Natalia Martinez}, \bibinfo{person}{Martin Bertran}, {and} \bibinfo{person}{Guillermo Sapiro}.} \bibinfo{year}{2020}\natexlab{}.
\newblock \showarticletitle{Minimax pareto fairness: A multi objective perspective}. In \bibinfo{booktitle}{\emph{International Conference on Machine Learning}}. PMLR, \bibinfo{pages}{6755--6764}.
\newblock


\bibitem[Narayanan et~al\mbox{.}(2011)]%
        {narayanan2011}
\bibfield{author}{\bibinfo{person}{Arvind Narayanan}, \bibinfo{person}{Elaine Shi}, {and} \bibinfo{person}{Benjamin I.~P. Rubinstein}.} \bibinfo{year}{2011}\natexlab{}.
\newblock \showarticletitle{Link prediction by de-anonymization: How We Won the Kaggle Social Network Challenge}. In \bibinfo{booktitle}{\emph{The 2011 International Joint Conference on Neural Networks}}. \bibinfo{pages}{1825--1834}.
\newblock
\urldef\tempurl%
\url{https://doi.org/10.1109/IJCNN.2011.6033446}
\showDOI{\tempurl}


\bibitem[Ng et~al\mbox{.}(2023)]%
        {google-research-identify-contrails-reduce-global-warming}
\bibfield{author}{\bibinfo{person}{Joe Ng}, \bibinfo{person}{Carl Elkin}, \bibinfo{person}{Aaron Sarna}, \bibinfo{person}{Walter Reade}, {and} \bibinfo{person}{Maggie Demkin}.} \bibinfo{year}{2023}\natexlab{}.
\newblock \bibinfo{title}{Google Research - Identify Contrails to Reduce Global Warming}.
\newblock
\newblock
\urldef\tempurl%
\url{https://kaggle.com/competitions/google-research-identify-contrails-reduce-global-warming}
\showURL{%
\tempurl}


\bibitem[On(2018)]%
        {bountyarticle}
\bibfield{author}{\bibinfo{person}{Amit Elazari~Bar On}.} \bibinfo{year}{2018}\natexlab{}.
\newblock \bibinfo{title}{We Need Bug Bounties for Bad Algorithms}.
\newblock
\newblock
\urldef\tempurl%
\url{https://www.vice.com/en/article/8xkyj3/we-need-bug-bounties-for-bad-algorithms}
\showURL{%
\tempurl}


\bibitem[Roelofs et~al\mbox{.}(2019)]%
        {roelofs2019meta}
\bibfield{author}{\bibinfo{person}{Rebecca Roelofs}, \bibinfo{person}{Vaishaal Shankar}, \bibinfo{person}{Benjamin Recht}, \bibinfo{person}{Sara Fridovich-Keil}, \bibinfo{person}{Moritz Hardt}, \bibinfo{person}{John Miller}, {and} \bibinfo{person}{Ludwig Schmidt}.} \bibinfo{year}{2019}\natexlab{}.
\newblock \showarticletitle{A meta-analysis of overfitting in machine learning}.
\newblock \bibinfo{journal}{\emph{Advances in Neural Information Processing Systems}}  \bibinfo{volume}{32} (\bibinfo{year}{2019}).
\newblock


\bibitem[Sven~Cattell(2023)]%
        {defcon}
\bibfield{author}{\bibinfo{person}{Austin~Carson Sven~Cattell, Rumman~Chowdhury}.} \bibinfo{year}{2023}\natexlab{}.
\newblock
\newblock
\urldef\tempurl%
\url{https://aivillage.org/generative%20red%20team/generative-red-team/}
\showURL{%
\tempurl}


\bibitem[Tosh and Hsu(2022)]%
        {tosh2022simple}
\bibfield{author}{\bibinfo{person}{Christopher~J Tosh} {and} \bibinfo{person}{Daniel Hsu}.} \bibinfo{year}{2022}\natexlab{}.
\newblock \showarticletitle{Simple and near-optimal algorithms for hidden stratification and multi-group learning}. In \bibinfo{booktitle}{\emph{International Conference on Machine Learning}}. PMLR, \bibinfo{pages}{21633--21657}.
\newblock


\bibitem[Tunguz et~al\mbox{.}(2023)]%
        {2023-kaggle-ai-report}
\bibfield{author}{\bibinfo{person}{Bojan Tunguz}, \bibinfo{person}{Dieter}, \bibinfo{person}{Karnika Kapoor}, \bibinfo{person}{Parul Pandey}, \bibinfo{person}{Paul Mooney}, \bibinfo{person}{Phil Culliton}, \bibinfo{person}{Rob Mulla}, \bibinfo{person}{Sanyam Bhutani}, {and} \bibinfo{person}{Will Cukierski}.} \bibinfo{year}{2023}\natexlab{}.
\newblock \bibinfo{title}{2023 Kaggle AI Report}.
\newblock
\newblock
\urldef\tempurl%
\url{https://kaggle.com/competitions/2023-kaggle-ai-report}
\showURL{%
\tempurl}


\end{thebibliography}

\onecolumn
\newpage
\label{Appendix}
\label{appendix-graphs}

\appendix

\section{Reproducibility}
Code to reproduce primary results and figures may be located at \url{https://anonymous.4open.science/r/Diversified-Ensembling-Reproducibility-52E0}

\section{Additional Plots}

\label{ap:extra-plots}

We provide additional plots used in the empirical analysis.

\begin{figure}[H]
\begin{center}
\includegraphics[scale=0.28]{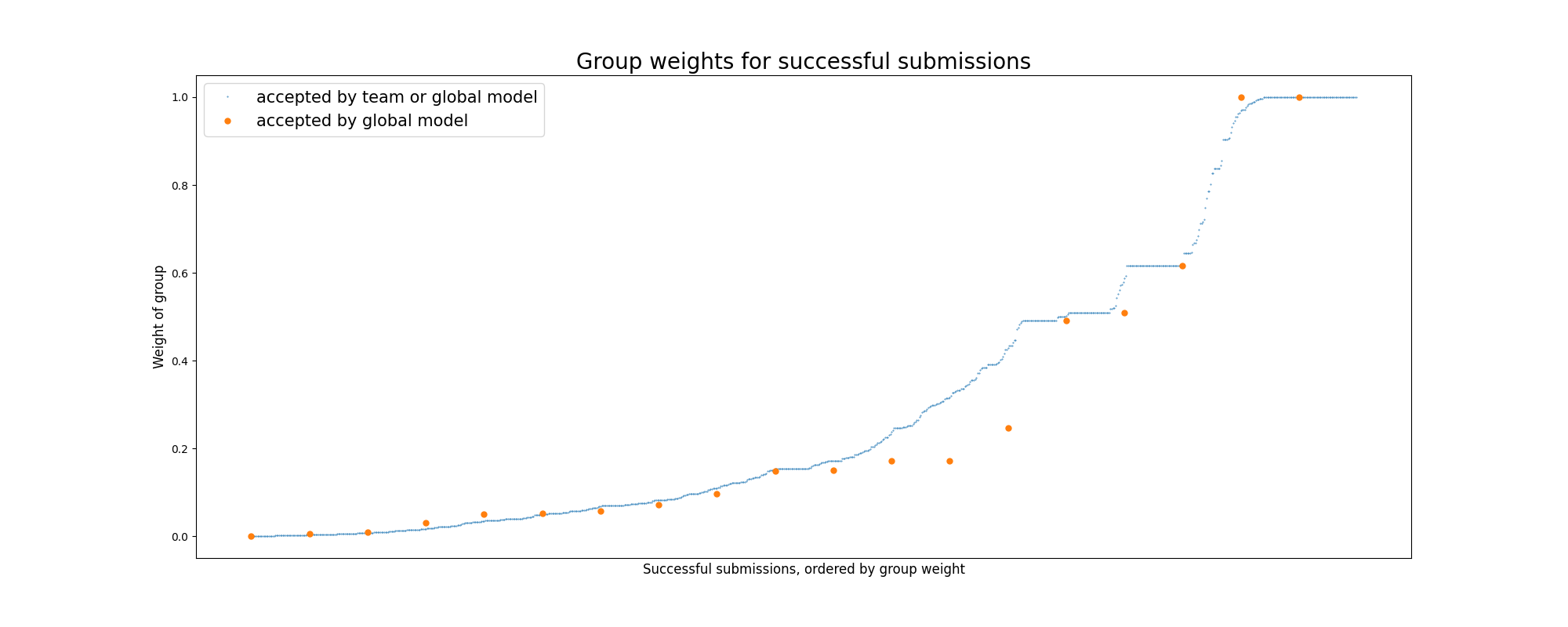}
\end{center}
\caption{Distribution of group weights for successful submissions. On the $y$-axis, weight of the group submitted, on the $x$-axis, individual submissions, as ordered by weight. The blue dots correspond to successful updates to either the global model or the team model, while orange dots correspond to submissions to the global model only.} 
\label{fig:acc-weight}
\end{figure}

\begin{figure}[H]
\begin{center}
\includegraphics[scale=0.31]{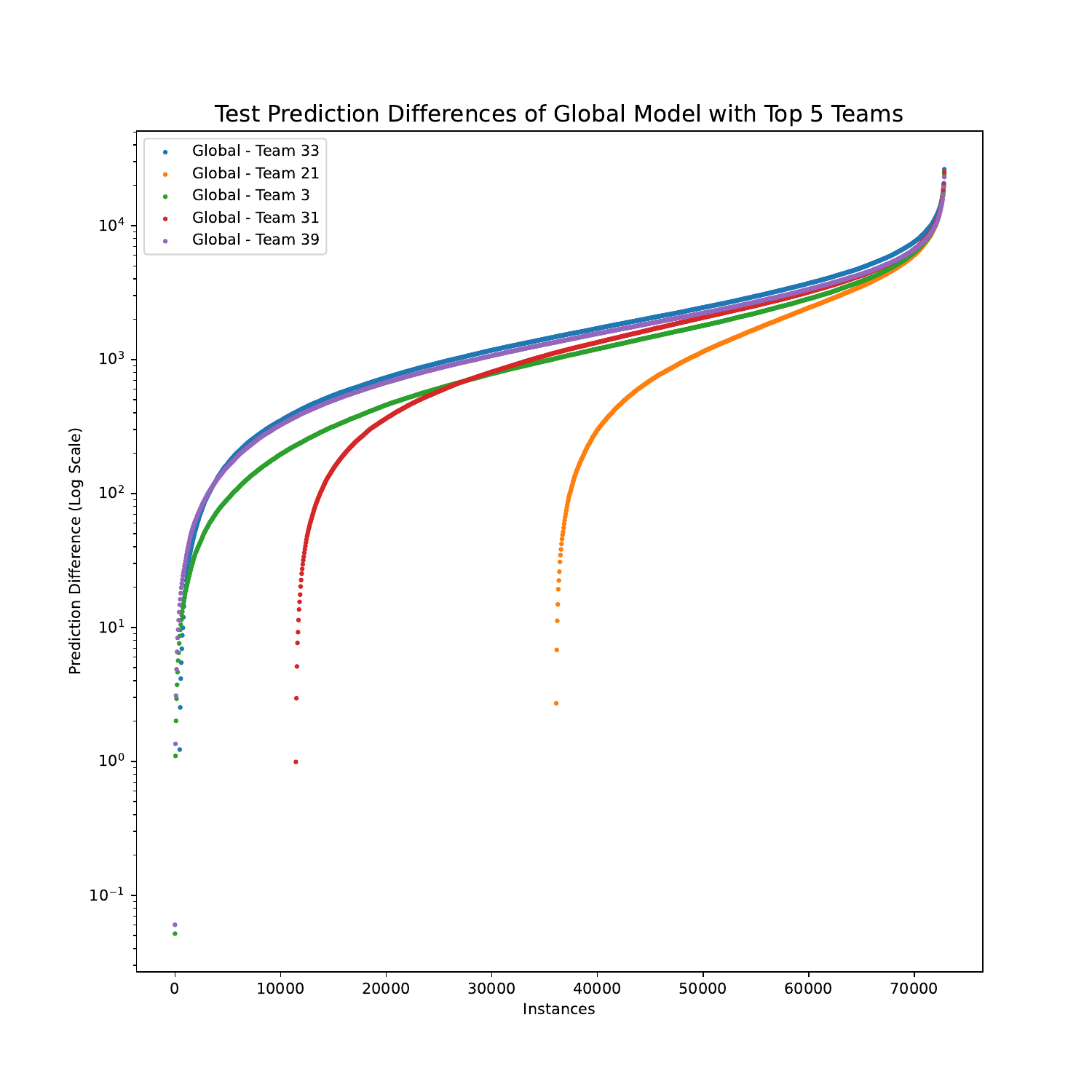}
\end{center}
\caption{Test prediction disagreement between the global model and top five local models. On the $x$-axis are individual datapoints, and on the $y$ axis the (absolute) difference in prediction of $y$ by the global model and an individual teams' model as measured on this point. The $y$-axis is log-scaled, and the disagreement is plotted for each of the top 5 teams. The datapoints are sorted on the $x$-axis so that the difference in prediction is monotonic. We see that while the top 5 teams had test loss that was relatively close to the global models' (see Table \ref{tab:results-table}), they make substantially different predictions for many points. }
\label{fig:top-5-diff}
\end{figure}

\begin{figure}[H]
\begin{tabular}{ccc}
  \includegraphics[width=.3\linewidth]{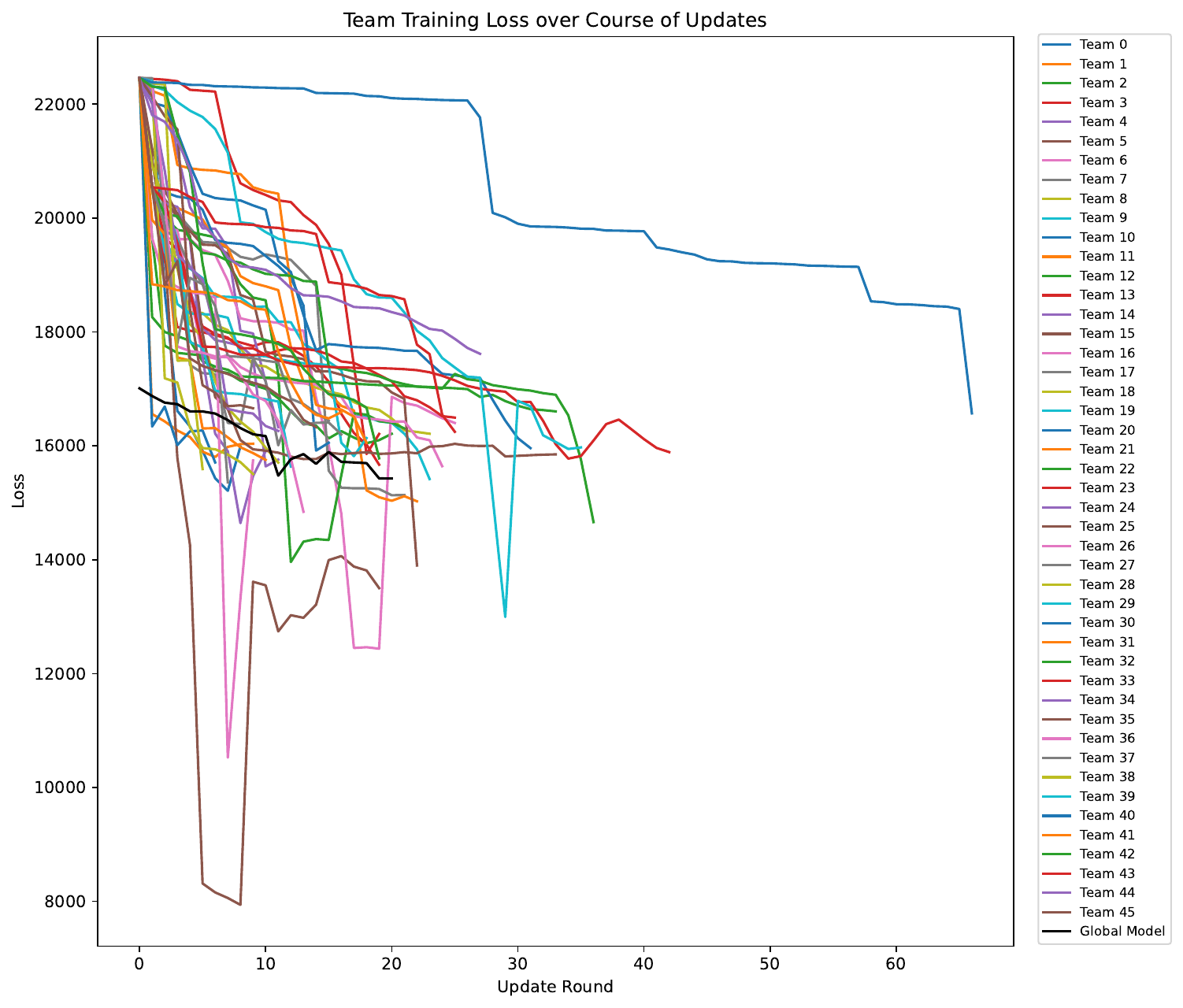} & \includegraphics[width=.3\linewidth]{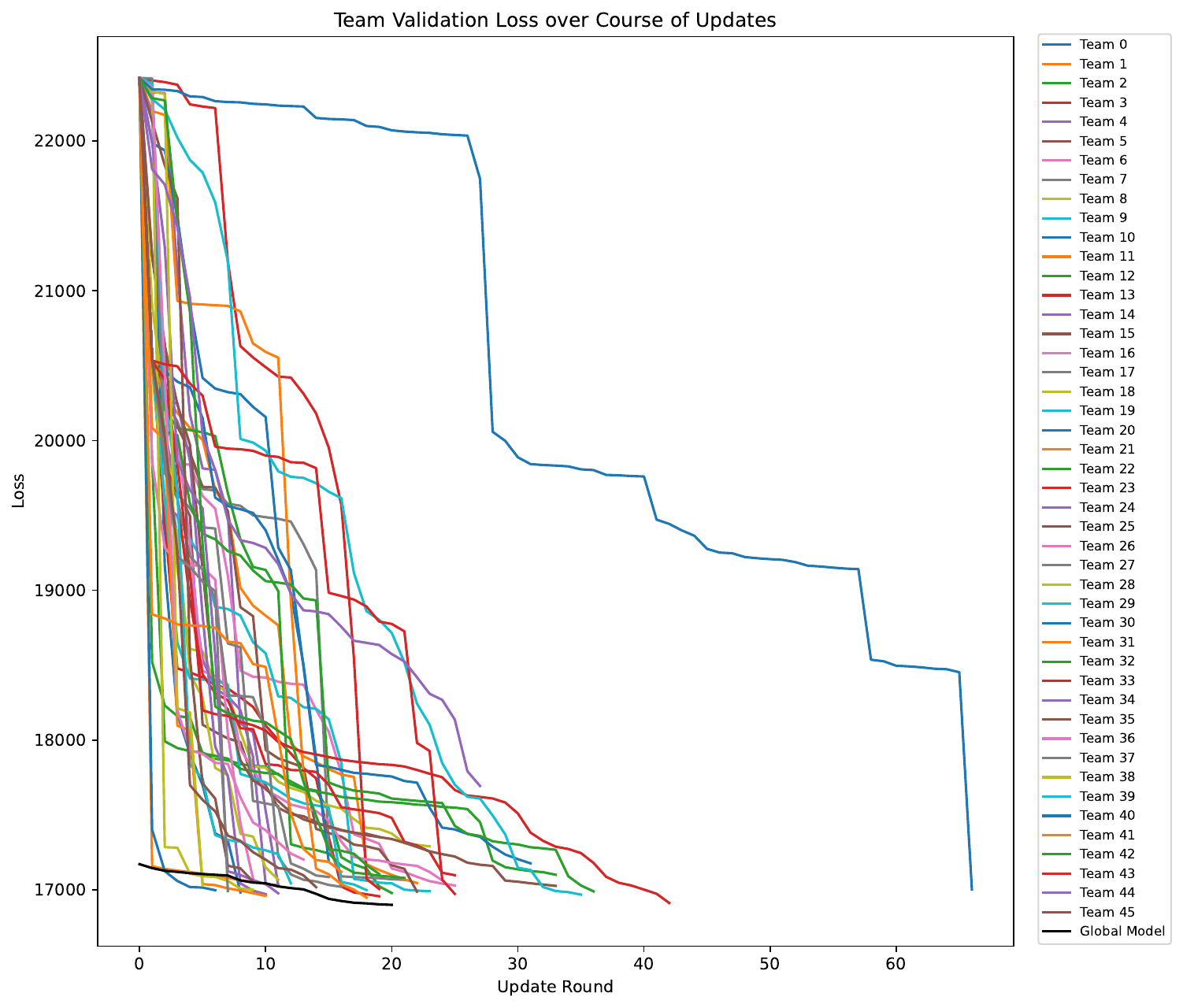} & \includegraphics[width=.3\linewidth]{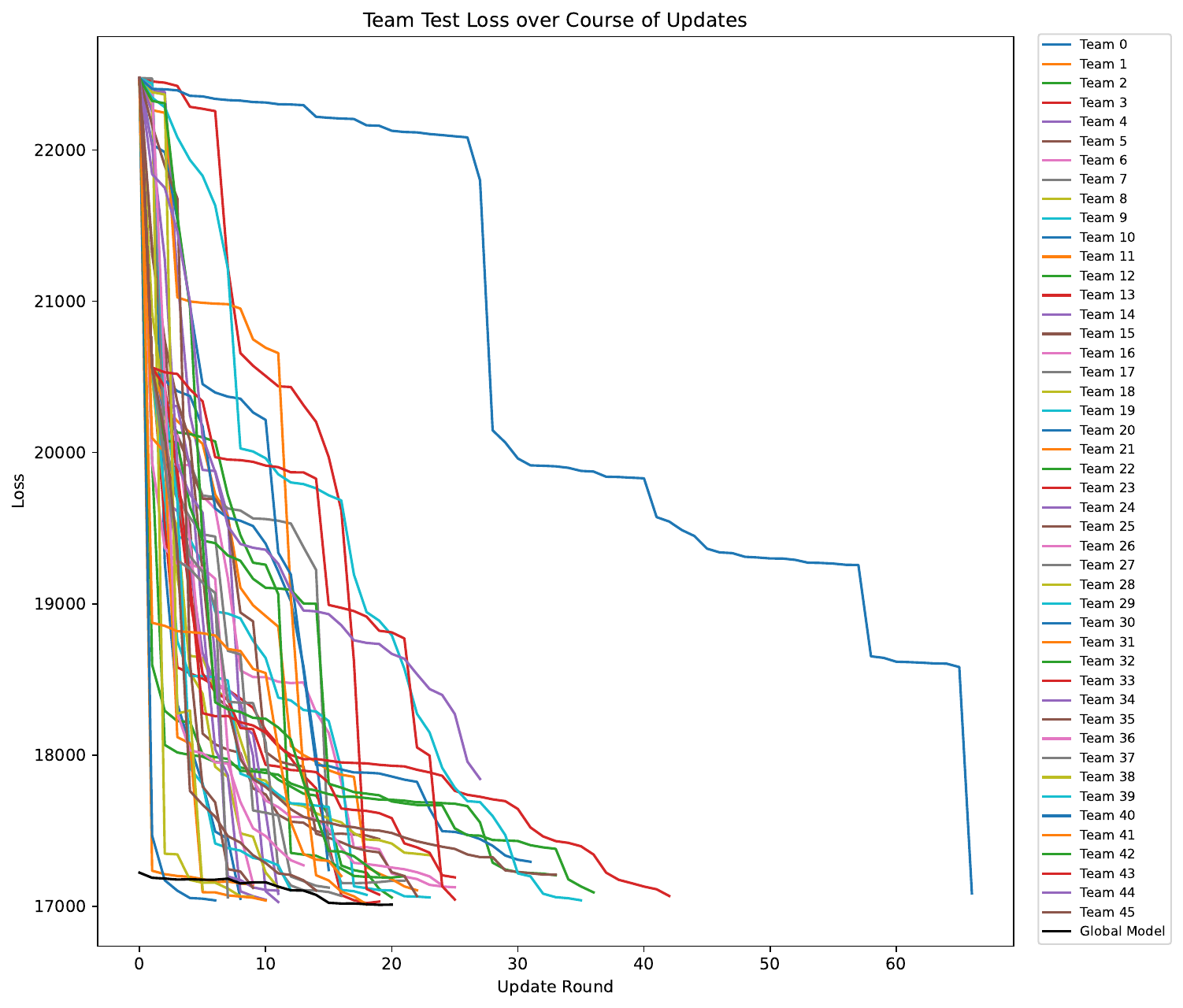}\\
(a) train & (b) validation & (c) test \\[6pt]
  \includegraphics[width=.3\linewidth]{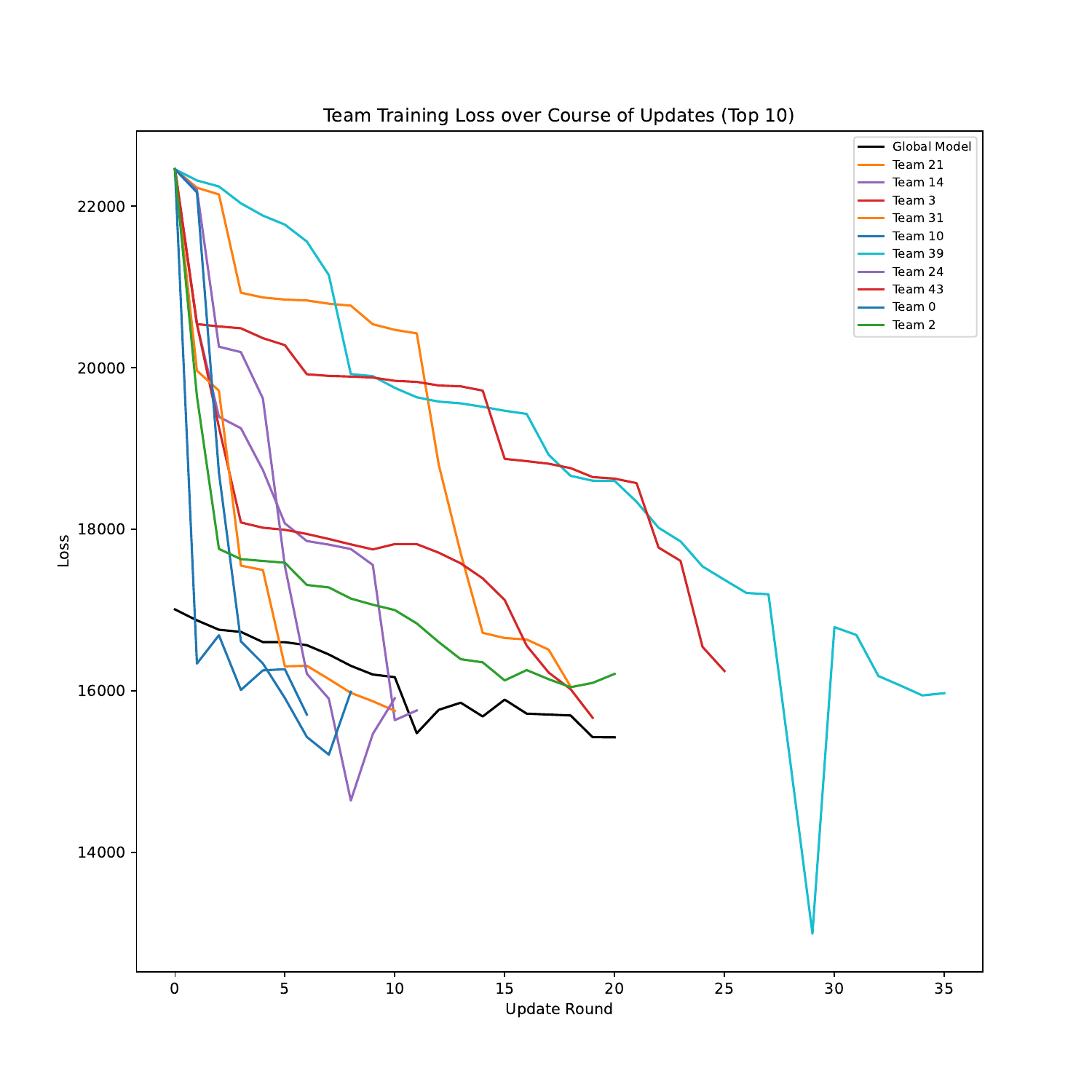} & \includegraphics[width=.3\linewidth]{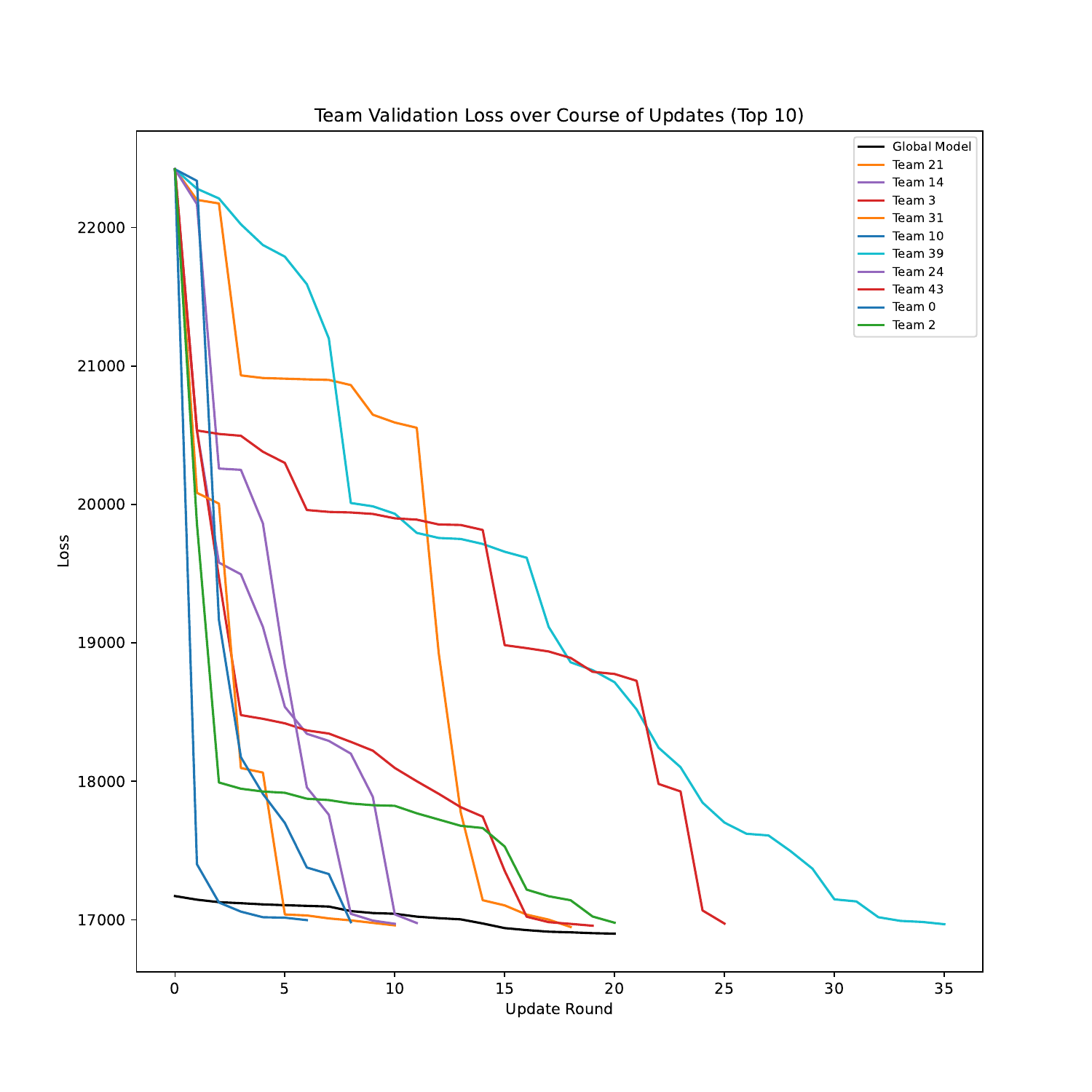} & \includegraphics[width=.3\linewidth]{temp_plots/team_test_errors_time_top_ten.pdf}\\
(a) train (top ten) & (b) validation (top ten) & (c) test (top ten) \\[6pt]
\end{tabular}
\caption{Error of global and teams' models measured at each accepted updated. Top row of subfigures consists of all teams while the bottom row displays the top ten teams and the global model. The $y$-axis is squared error, the $x$-axis is the round of accepted submission.}
\end{figure}

\begin{table}[]
    \centering
\begin{tabular}{llllr}
        Team & Training Loss & Validation Loss & Test Loss &  Number of Updates \\
\hline
Global Model &      15424.70 &        16900.12 &  17010.49 &                 20 \\
     Team 21 &      16052.41 &        16948.26 &  17012.32 &                 18 \\
     Team 14 &      15757.40 &        16977.12 &  17028.13 &                 11 \\
      Team 3 &      15667.19 &        16957.37 &  17030.65 &                 19 \\
     Team 31 &      15752.59 &        16959.99 &  17037.82 &                 10 \\
     Team 10 &      15706.57 &        16997.82 &  17038.68 &                  6 \\
     Team 39 &      15971.36 &        16968.34 &  17039.29 &                 35 \\
     Team 24 &      15907.08 &        16971.54 &  17043.18 &                 10 \\
     Team 43 &      16245.32 &        16972.86 &  17044.38 &                 25 \\
      Team 0 &      15987.90 &        16982.91 &  17049.55 &                  8 \\
      Team 2 &      16210.24 &        16979.03 &  17058.09 &                 20 \\

\end{tabular}
\caption{Final leaderboard of top ten participating teams sorted by increasing test error with number of updates made to each model. 
}
    \label{tab:results-table}
\end{table}

\section{Data Features}
\label{ap:data-features}

In the following Table \ref{table:pums-features}, the full list of features used in the income prediction task students were competing over are listed. The exact categories within each feature and the formal descriptions of categories listed here are from the PUMS data dictionary (\cite{pums-dictionary}). 
\begin{table}[H]
\begin{centering}
\begin{tabular}{c c}
Feature & Feature Description\\
\hline  \\
ST & State and territory codes \\
AGEP & Age \\
CIT & Citizen status \\
COW & Class of worker \\
DDRS & Self care difficulty \\
DEAR & Hearing difficulty \\
DEYE & Vision difficulty \\
DOUT & Independent living difficulty \\
DRAT & Veteran service connected disability rating \\
DREM & Cognitive difficulty \\
ENG & Ability to speak English \\
FER & Gave birth to child within the past 12 months \\
JWTRNS & Means of transportation to work \\
LANX & Language other than English spoken at home \\
MAR & Marital status \\
MIL & Military service \\
SCHL & Educational attainment \\
SEX & Binary sex \\
WKHP & Usual hours worked per week past 12 months \\
OCCP & Occupation code \\
RAC1P & Race code \\
\end{tabular}
\caption{Features from ACS PUMS survey used for the competitions' income prediction task.}
\label{table:pums-features}
\end{centering}
\end{table}

\section{Additional Observations}
\label{ap:emp}

As our competitors were students and not machine learning experts, some of the choices they made are perhaps counter-intuitive. We list some of these choices here.

\textbf{Most teams never attempted to make global improvements to the model:} Students were primarily graded in terms of their improvements to their team model, as opposed to the global model, as we anticipated that updates to the global model would become rapidly difficult to find for many students. Since their team models were initially trained as decision tree stumps with marginally more predictive power than predicting the mean label, a valid strategy would have been to simply improve this model over the entire dataset. However, only 15 of the 45 teams attempted updates over the whole dataset. 

\textbf{Teams may not have carefully examined their group functions:} For instance, out of the 222 updates which assigned $g$ to the entire dataset, 42 were complex logical combinations of many features, which also happened to include a disjunction of, e.g., binarized sex labels, and hence the entire dataset. This could have been due to mistakes manually conditioning over group membership (e.g. missing parentheses in the logical operations or a swap of an OR and an AND). Or it might be due to automated processes that teams used to identify groups, leading to a complex-looking group that can be much simplified. In either case, it might indicate that teams were not carefully engaging with what they were submitting. 

Similarly, of the 95 submissions which considered the binary feature for presence of a hearing disability (which were submitted by 22 distinct teams), \textit{all of them} specified groups considered exclusively the hearing loss feature marginalized on \textit{not} having hearing loss. The presence of a disability might have predictive power that conceivably could be ignored by a larger model, since they make up a minority population (about 2.3 percent for hearing loss). However, predicating on \textit{not} having hearing loss amounts to considering the majority of the dataset, and one might postulate that this is not a meaningful form of specialization. Students did not discuss this choice in their write up. However, the way ACS PUMS categorizes disability is a bit unusual: they are categorical variables with value 1 being presence of the disability, and value 2 being absence, rather than a binary encoding. Thus without carefully checking the categorization of values, students might have misunderstood the meaning of values, though we have no way of verifying this.

We expect that capping submissions per day would somewhat mitigate this, as it incentivizes teams to be relatively confident in their submission and hence inspect it closely. 

\textbf{Teams conflated societal inequity with disparate model performance} As discussed in Section \ref{sec:analysis}, teams often chose to specialize on groups who face systemic income inequity. While the performance of machine learning models may very well be disparate across such groups, it is not necessarily always the case. For instance, if there is a strong signal in the training data that income inequity between two sufficiently large groups exists, a model trained on the entire dataset will likely use this signal. The bigger issue here is often in the prediction task itself, the selection of features used for the task, or under-representation in the dataset. Finding disadvantaged groups where the existing training data \textit{can} be leveraged to improve performance in a way that a model trained over the entire dataset wouldn't capture is potentially nontrivial, especially without access to external data sources. 

\section{Platform Deep Dive}
\label{ap:deep-dive}
In this section, we expand on deeper technical details of the platform introduced in Section \ref{Platform Design}. The end goal of this section is to provide the reader an explanation for all steps listed in Figure \ref{fig:infrastructure-larger} in an end-to-end platform loop from submitting $(g,h)$-pairs to receiving feedback.

\subsubsection{Submitting and testing $(g,h)$ pairs} 
As previously mentioned in Section \ref{Platform Design}, the method for submitting a $(g,h)$-pair for update is through a pull request on the competition's repository. Once a participant has trained a $(g,h)$-pair of models, they upload them to a cloud storage space such as Google Drive or AWS S3 with unauthenticated accessibility. To submit their pair, participants create a pull request on the repository hosting the competition by inserting the links to their models in the file \textit{/competitors/request.yaml} next to the variables \textit{g\_url} and \textit{h\_url}. The platform infrastructure will download and test this pair and provide feedback to the participant through a comment on their pull request.

\subsection{Repository permissions and GitHub Actions integration}
Repository permissions are set such that all members with access are able to instantiate a pull request but only workspace administrators may push to the repository, following GitHub's standard structure for making updates to an open-source repository. In order to automatically test updates from participants in a competition, the platform utilizes GitHub Actions to spawn a \textit{workflow} script anytime a user creates a pull request. Our specific workflow first checks for any file changes made outside of the \textit{/competitors/request.yaml} file used for submissions, and runs the backend procedure for testing $(g,h)$-pair updates on models.

\subsubsection{Docker Integration}
An important implementation detail not discussed in Section \ref{Platform Design} is the usage of Docker in our platform. In order to protect the host machine and standardize the operating environment, the server back-end computations are run inside docker containers. The \textit{repository} container maintains the file system for teams models, interacts with \textit{GitHub Actions} when participants submit requests, and pushes information changes to GitHub when updates are made. The \textit{security} container's sole purpose is to run forward passes of submitted models to indicate to the repository container if the model is safe or potentially unsafe. The security container runs a primitive security check on models, scanning for malicious keywords and opcodes. The security container is not connected to the internet in order to reduce potential risks to the host machine and competition results. 

\subsubsection{End-to-End Submission Workflow}
\begin{figure}[H]
\begin{center}
\includegraphics[width = .9\linewidth]{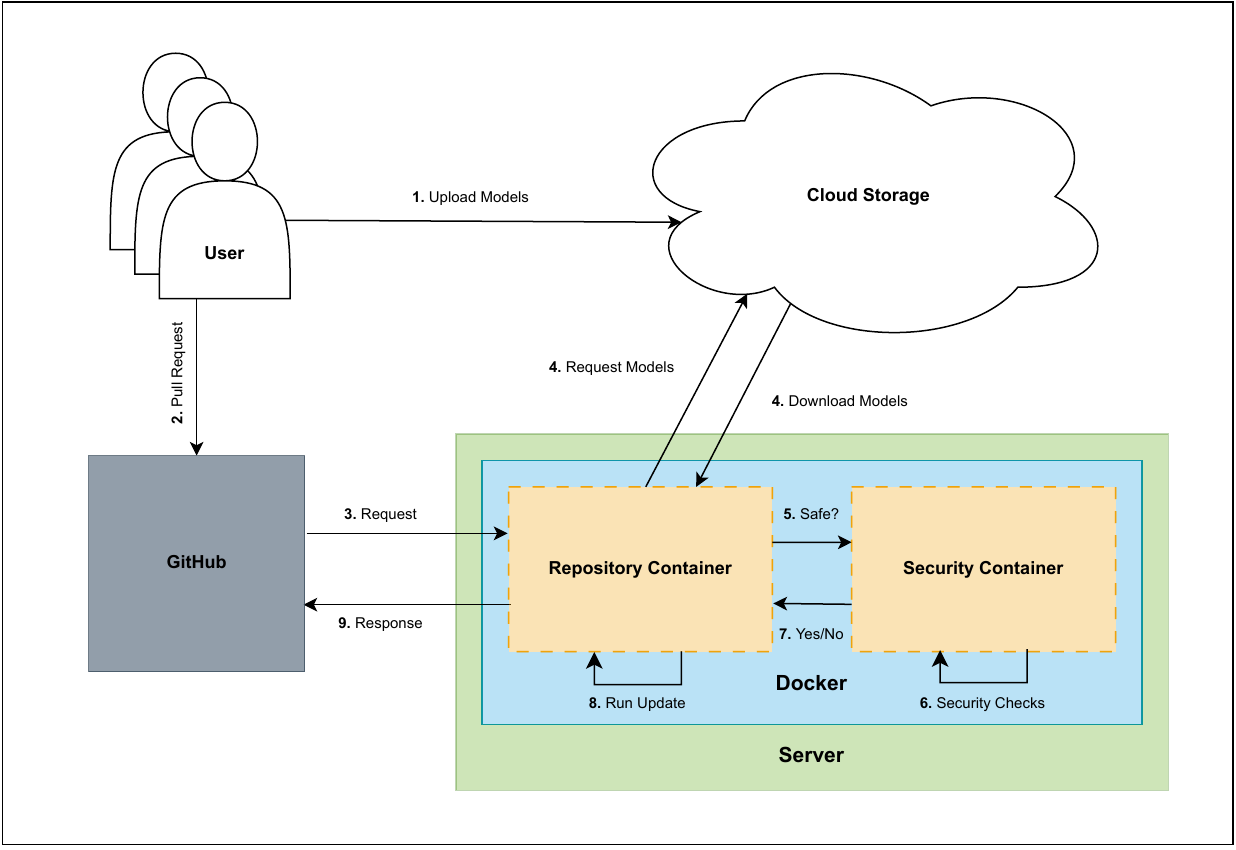}
\end{center}
\caption{Protocol for (g,h) pair submissions from participants to a competition repository.}
\label{fig:infrastructure-larger}
\end{figure}

Using Figure \ref{fig:infrastructure-larger} as reference, we expand on the steps in the platform feedback loop for submitting pairs to a competition repository.  
\begin{enumerate}
    \item User saves and uploads trained g,h models to cloud storage with publicly accessible URLs.
    \item User submits pull request to GitHub repository editing \textit{competitors/request.yaml} file to include g,h model URLS.
    \item GitHub Actions notifies Repository Container of pull request from user.
    \item Repository Container downloads changed files and downloads models from URLs.
    \item Repository Container notifies Security Container of new models and requests forward pass.
    \item Security Container runs primitive security checks on models and ensures outputs are correctly formatted.
    \item Security Container informs Repository Container of safe/unsafe models
    \item If models are safe, Repository Container loads team and global models into memory and attempts updates. If unsafe, skip step.
    \item Repository Container closes/makes comments on Pull Request and deletes branch. If updates are accepted, changes are pushed to GitHub.
\end{enumerate}

\section{Consent Form and IRB Exemption}
\label{ap:consent}

This study was deemed IRB Exempt under category 1; below is the consent form that students signed for the purposes of this study.

\textbf{Project Title:} Bias Bounties Analysis

\subsection{Summary and Purpose of the study}
This research is being conducted by (omitted for anonymization) at (omitted for anonymization). The purpose of this research is to understand the heuristic approaches and methodologies used to isolate regions where machine learning models perform poorly. Your participation will not require any work additional to your work in the course and will have no risks or benefits to you directly, but may lead to better understanding of how to improve machine learning models in the future. Your participation is voluntary and will not affect your grade in the course. Your work will be de-identified, and as there is no personal information present in your machine learning models, re-identification is unlikely. The data will be stored, and could be used for future research.

\subsection{Procedures}

We will analyze your machine learning models submitted as part of the final course project for (omitted for anonymization). We may also look at your (de-identified) source code and final project write-ups to make qualitative conclusions about your approaches.

\subsection{Confidentiality}

While personally identifiable information in this context is extremely minimal, your names will not be associated with your models in the context of the research study, and no information about individuals will be accessible from the published work.

\subsection{Benefit and Risks to Participants}

There are no specific benefits or risks to you in particular if you join this study. Whether or not you do participate will not affect your grade in this class.

\subsection{Right to Withdraw}

Your participation in this research is completely voluntary. You may refuse to participate or withdraw at any time before the end of the semester.

\subsection{Contact information for questions}

(Omitted for anonymization)

\section{Project Description}
\label{ap:proj-desc}

Here we include the project description that students received at the outset of the assignment. 

\noindent \textbf{Bias Bounties Project}
\textit{Due date: April 21st, 2023}

\section*{Task}
In this project, your team will act as ``Bias Bounty Hunters" for a regression model trained with sole intent of minimizing empirical loss (\textit{root mean squared error}). The model accepts US Census data from the Folktables package with predefined features and predicts the annual income of an individual. Your team is tasked with finding \textit{certificates of suboptimality} in order to reduce the correctable bias in the model. These certificates take the form $(g,h)$, where $g$ is a group indicator function, $h$ is a regression model, and $h$ performs strictly better on $g$ than the current model. 

\section*{Class Semantics + Notation}
Since we expect the communal model to quickly converge to approximate Bayes-Optimal, each team will also have a \textit{team model}. We will refer to the communal predictive model as the \textit{global model} and each team's model as their \textit{private model}. Both of these models are Pointer Decision Lists which follow the update procedure we discussed in class. Each time a team submits a potential update to the global model, the update will also be attempted on the team's private model. This is to incentivize all teams to continue searching for updates since you will be graded on your private model's accuracy (and given extra credit based your teams change to the global model's accuracy).

\section*{Submitting Certificates of Suboptimality}
In order to submit potential $(g,h)$ pair updates, you will create a pull request on the GitHub repository with the file /competitors/request.yaml altered to reflect the URLs of your models. If any other file is altered, the pull request will immediately be denied. For explicit, detailed instructions, please refer to the GitHub README file. 

\section*{Deliverables}
Each team will deliver one report with the information in the template populated. The template contains formatting for your report as well as descriptions on how to fill out each section. Any amount of information which goes over the section's word limit will not be used when grading. While a Bias Bounty Competition is generally focused around individual teams' updates to the global model's accuracy, this course is not expecting to produce expert ML practitioners. Your private model's accuracy will be included in the overall grading scheme but we are much more interested in \textit{how} your team attempted to find updates. Furthermore, while we want your team to find working methods, \textbf{we also care about what methods you tried that did not work since this is a heavily understudied field.} You should document your methods as you work on the project!

\section*{Cheating Policy}
The University of Pennsylvania Code of Academic Integrity is in effect and attempted malicious actions will not be taken lightly. If you or a member of your team are caught cheating or acting maliciously, your team will be removed from the project and will receive a score of 0.

\end{document}